%% file: neurips_2026.tex
\documentclass{article}

\usepackage[main, final]{neurips_2026}

\usepackage[utf8]{inputenc} 
\usepackage[T1]{fontenc}    
\usepackage{hyperref}       
\usepackage{url}            
\usepackage{booktabs}       
\usepackage{amsfonts}       
\usepackage{nicefrac}       
\usepackage{microtype}      
\usepackage[table]{xcolor}  

\usepackage{graphicx}
\usepackage{subcaption}
\usepackage{multirow}
\usepackage{makecell}
\usepackage{amsmath}
\usepackage{amssymb}
\usepackage{float}
\usepackage{tabularx}
\usepackage{rotating}
\usepackage{placeins}
\usepackage{afterpage}
\newcolumntype{Y}{>{\raggedright\arraybackslash}X}

\DeclareMathOperator{\softmax}{softmax}

\title{Revisiting Continual Learning in Language Models:\\ A Three-Interface Diagnosis of Stability and Plasticity}

\author{%
  Keuntae Kim$^{1*}$ \quad
  Beomseok Lee$^{2*}$ \quad
  Hyunwoo Kim$^{3*}$ \quad
  Yong Suk Choi$^{1\dagger}$\\
  {\normalfont $^{1}$Hanyang University \quad $^{2}$LG Electronics \quad $^{3}$KT Corporation}\\
  {\normalfont\texttt{ktkpv94@hanyang.ac.kr}}
}

\begin{document}

\maketitle

\begingroup
\renewcommand{\thefootnote}{}
\footnotetext{$^{*}$Equal contribution. \quad $^{\dagger}$Corresponding author.}
\endgroup

\input{sections/0.Abstract}
\input{sections/1.Introduction}
\input{sections/2.Background}

\input{sections/3.Method}
\input{sections/4.Experimental_Setup}
\input{sections/5.Results}

\input{sections/6.Analysis}
\input{sections/7.Conclusion}

\begin{ack}
This work was supported by the Institute of Information \& Communications Technology Planning \& Evaluation (IITP) grant (No.~RS-2025-25422680, No.~RS-2020-II201373), and the National Research Foundation of Korea (NRF) grant (No.~RS-2025-00520618) funded by the Korean Government (MSIT).
\end{ack}

\bibliography{references}
\bibliographystyle{plainnat}

\newpage
\appendix
\input{sections/8.Appendix}

\newpage
\input{checklist.tex}

\end{document}

%% file: sections/0.Abstract.tex
\begin{abstract}
Continual learning in language models is commonly studied in the incremental learning (IL) setting, where a model learns a sequence of tasks and is evaluated through its final classifier or language-model (LM) head. This interface, however, does not indicate where forgetting occurs: a drop in accuracy may reflect representation degradation in the backbone, readout mismatch in the head, or both. We therefore argue that backward transfer (BWT) and forward transfer (FWT) should be interpreted as interface-dependent quantities rather than as single model-level properties. We study three evaluation interfaces: the trained head, which is updated sequentially with the backbone; a probe newly trained on frozen backbone features; and a classifier-free backbone diagnostic. They measure deployed performance, recoverability under a new readout, and representation separability, respectively. Across class-incremental experiments with encoder and decoder language backbones, plus task-incremental sanity checks, conclusions about forgetting and plasticity change with the interface: the trained head often indicates severe forgetting, the probe near-complete recoverability, and the backbone diagnostic intermediate representation drift with a consistent negative relation between stability and plasticity. This relation holds for clustering-free diagnostics such as nearest-class-mean accuracy and across five clustering algorithms and three metrics. Finally, Just LM-Head Tuning (JLT), a head-only intervention on a fixed incrementally trained backbone, recovers 96--99\% of the gap to joint training for replay- and distillation-based methods but only about three quarters for plain sequential fine-tuning, suggesting that many failures attributed to catastrophic forgetting reflect head--backbone mismatch plus partial representation drift. Our results call for reporting where forgetting is measured, not only how much. Code is available at \url{https://github.com/kkt94/RevisitingCL}.

\end{abstract}

%% file: sections/1.Introduction.tex
\section{Introduction}
\label{sec:intro}
Continual learning studies how a model can keep acquiring new knowledge while retaining what it has already learned \citep{parisi2019continual}. A widely studied setting is incremental learning (IL), in which a model is trained on a sequence of tasks and evaluated on all tasks seen so far \citep{polikar2001learn++, kemker2017fearnet}. In language modeling and NLP, this problem is especially important because pretrained language models (PLMs) are often adapted to new domains or task streams through sequential fine-tuning. A persistent challenge is catastrophic forgetting: performance on earlier tasks can degrade sharply after the model is updated on later tasks \citep{french1999catastrophic, li2019learn, hu2019overcoming, kaushik2021understanding, van2024continual}. Many IL methods therefore focus on memory stability through replay, regularization, or distillation \citep{kirkpatrick2017overcoming, wang-etal-2023-rehearsal, yang2024recent}.

However, standard IL evaluation usually observes the model through the final classifier or language-model (LM) head. This interface is useful for measuring deployed task accuracy, but it can also mix together two different phenomena: degradation of the backbone representation and failure of the prediction head to read out information that is still present in the backbone. This distinction matters for the \textit{Stability--Plasticity Dilemma}~\citep{abraham2005memory, mermillod2013stability, wu2021striking, araujo2022entropy}. If forgetting is mainly a head-level readout problem, then low head-level accuracy does not necessarily imply that the backbone has irreversibly lost the task. Conversely, if a newly trained probe compensates for weak or shifted representations, it may understate representational forgetting.

\begin{figure}
\begin{center}
\includegraphics[width=1.0\linewidth]{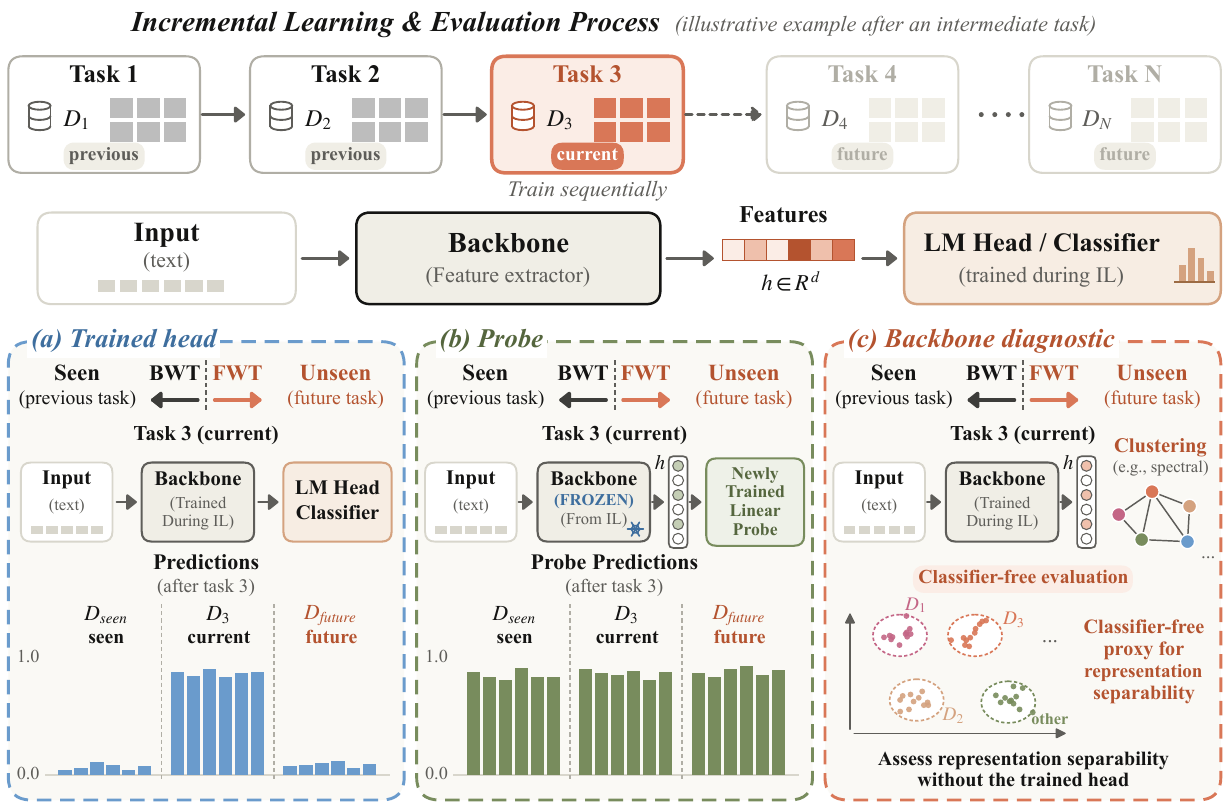}
\end{center}
\caption{Overview of evaluating BWT and operational FWT at three interfaces in discriminative backbones. We measure transfer through (a) the trained head, (b) a probe, and (c) the backbone diagnostic, a classifier-free proxy for representation separability. The backbone diagnostic does not replace task accuracy; it evaluates separability without using the trained head. In generative backbones, the last-token hidden state of the prompt, from which the label is generated, replaces the [CLS] feature, and the LM head serves as the trained head. Details are in Appendix \ref{Appendix_A}.}
\label{fig:architecture}
\end{figure} 

Recent studies suggest that the backbone can retain more information than head-level accuracy indicates, and that much of the observed degradation can arise in the classifier or LM head \citep{davari2022probing, seq*, zheng2025spurious}. Other work studies Forward Transfer (FWT) and argues that less forgetting can provide a useful inductive bias for future tasks \citep{chen2023forgetting, zheng2025towards}. These findings raise a more basic measurement question: when an incremental learner fails on old tasks, where is forgetting being measured?

We address this question by revisiting Backward Transfer (BWT) and FWT under three evaluation interfaces: the trained head (the classifier or LM head updated sequentially with the backbone), a probe (a linear classifier newly trained on frozen backbone features), and the backbone diagnostic (classifier-free separability of backbone features). The main contribution is not a new metric that replaces accuracy. It is a diagnostic decomposition showing that standard IL metrics conflate three distinct estimands: deployed head behavior, recoverable information under a new readout, and classifier-free representation separability. Since the backbone has no standalone prediction interface, we approximate changes in backbone separability using clustering-based and clustering-free diagnostics.

Our analysis shows that the apparent relationship between BWT and FWT changes substantially across these interfaces. Head-level metrics often overstate forgetting because the head is itself unstable under sequential updates, while the probe can understate forgetting because a newly trained classifier can compensate for shifted or weakened representations. The backbone diagnostic suggests intermediate representation drift and a consistent stability--plasticity pattern across clustering-free diagnostics and clustering choices.

Motivated by this gap between interfaces, we study Just LM-Head Tuning (JLT), a simple head-only adaptation procedure. JLT is not a new continual learning algorithm but a diagnostic intervention: by re-optimizing only the LM head on top of a fixed, incrementally trained backbone, we estimate how much performance can be recovered by correcting head--backbone mismatch. Because the head may be adapted before each task is evaluated, JLT scores should be read as recoverable performance, not as standard class-incremental learning (CIL) accuracy.

We evaluate this diagnostic view across five benchmark groups spanning seven datasets, encoder and decoder language backbones, multiple IL methods in CIL, and a discriminative-backbone sanity check in task-incremental learning (TIL). Concretely, we formalize BWT and an operational FWT (plasticity-gain) measure as interface-dependent quantities: the same formula answers different questions when measured through the trained head, a probe, or the backbone diagnostic. On this basis, we provide a three-interface diagnostic framework that separates deployed performance, readout recoverability, and classifier-free representation separability. Empirically, we show that these interfaces lead to qualitatively different conclusions, that the backbone-level stability--plasticity pattern is consistent across clustering-free diagnostics and clustering choices, and that JLT can be used as a head-only intervention to estimate how much apparent forgetting is recoverable through readout re-alignment and how much remains as a representation-level residual. Unlike prior probing studies, which examine stability only~\citep{davari2022probing, seq*, zheng2025spurious}, this framing measures representation-level plasticity and compares the BWT--FWT relation across interfaces (Appendix Table~\ref{tab:prior_comparison}).

%% file: sections/2.Background.tex
\section{Metrics are interface-dependent}
\subsection{Problem setup}
\label{sec:2.1}
We consider an incremental learner that receives a sequence of tasks \(D = \{D_1, D_2, \cdots, D_T\}\), where each task \(D_t = \{(x_t^i, y_t^i)\}_{i=1}^{N_t}\) contains inputs \(x_t^i \in X_t\) and labels \(y_t^i \in Y_t\). The learner updates the same model sequentially from task 1 to task \(T\).

The two most common IL scenarios are CIL and TIL. In CIL, label sets from different tasks do not overlap, i.e., \(Y_i \cap Y_j = \emptyset\) for \(i \neq j\), and the model must predict among all classes observed so far. In TIL, the task identity is available at evaluation time, so prediction is restricted to the classes of the corresponding task. Because CIL requires a shared prediction space across tasks, it is usually more challenging than TIL \citep{tao2023can}. We focus on CIL in the main results and use TIL as a sanity check in Appendix~\ref{appendix:D}.

\subsection{Evaluation metrics for IL}
\label{sec:2.2}

\textbf{BWT} \citep{lopez2017gradient, ebrahimi2018uncertainty, ebrahimi2020adversarial} is a representative IL metric for \textit{memory stability}. It compares performance on a previous task immediately after that task is learned with performance on the same task after the full task sequence has been learned:
\begin{equation}
\label{equation:BWT}
BWT = \frac{1}{T-1} \sum_{i=1}^{T-1} \left( a_{T,i} - a_{i,i} \right)
\end{equation}
where $T$ is the last task, $a_{T,i}$ is the score on the \(i\)-th task after training up to the \(T\)-th task, and $a_{i,i}$ is the score on the \(i\)-th task immediately after training that task. The score \(a\) can be instantiated at three evaluation interfaces: the trained head, a probe, or the backbone diagnostic (Section~\ref{sec:estimands}). We denote the resulting metrics BWT-head, BWT-probe, and BWT-backbone, and likewise for FWT. These instantiations answer different questions and should not be treated as interchangeable.

\textbf{FWT} measures \textit{learning plasticity}: how much the model improves on a task after actually training on it, relative to its state before seeing that task. This is harder to measure than BWT because a standard classifier often has no output units for unseen classes before the task is introduced. Prior work therefore often uses random accuracy as a proxy \citep{ke2020continual, wolczyk2021continual, ke2022continual, wang2024comprehensive} or trains a separate classifier to estimate performance on the next task \citep{chen2023forgetting}. In this work, we use the following operational form and instantiate it separately for each evaluation interface:
\begin{equation}
\label{equation:FWT}
FWT = \frac{1}{T-1} \sum_{i=2}^{T} \left( a_{i,i} - a_{i-1,i} \right)
\end{equation}
where $T$ is the last task, $a_{i,i}$ is the score on the \(i\)-th task immediately after learning the \(i\)-th task, and $a_{i-1,i}$ is the score on the \(i\)-th task before training on it. For the trained head, this captures the behavior of the deployed prediction interface. For the probe, it captures how much a newly trained classifier can recover from frozen representations. For the backbone diagnostic, it approximates changes in backbone separability without any trained readout.

We use the term FWT in the operational sense common in continual learning evaluation, but the quantity in Eq.~(\ref{equation:FWT}) should be read as a task-exposure plasticity gain. It measures how much the model improves on task \(i\) after training on task \(i\), relative to the state immediately before seeing that task. This differs from purely zero-shot forward transfer. We keep the term FWT for consistency with IL literature, but throughout the paper we interpret it as an interface-specific plasticity-gain measure.

\subsection{Three estimands and a head-only intervention}
\label{sec:estimands}
BWT and operational FWT are formulas over scores \(a_{t,i}\), but the score itself is defined by an evaluation interface. Let \(m_t=(f_t,g_t)\) denote the model state after training up to task \(t\), where \(f_t\) is the backbone and \(g_t\) is the sequentially trained prediction head, which we call the trained head. An evaluation interface \(e\) specifies which components are frozen, which readout is used or fitted, what auxiliary data are allowed for fitting, and which scoring rule is applied. It induces an interface-specific score
\[
a^e_{t,i}=S_e(m_t;D_i^{\mathrm{tr}},D_i^{\mathrm{te}}),
\]
where \(D_i^{\mathrm{tr}}\) and \(D_i^{\mathrm{te}}\) denote the task split available to the evaluation protocol. BWT and operational FWT are therefore not properties of \(m_t\) alone, but of the pair \((m_t,e)\):
\[
\mathrm{BWT}^{e}=\frac{1}{T-1}\sum_{i=1}^{T-1}
\left(a^{e}_{T,i}-a^{e}_{i,i}\right),
\]
\[
\mathrm{FWT}^{e}_{\mathrm{op}}=\frac{1}{T-1}\sum_{i=2}^{T}
\left(a^{e}_{i,i}-a^{e}_{i-1,i}\right).
\]
Reporting BWT or operational FWT without specifying \(e\) is therefore an underspecified measurement. Different interfaces can induce different signs and magnitudes because they instantiate different scoring functionals over the same model trajectory.

The trained head measures deployed behavior, the probe measures recoverability under a newly optimized readout, and the backbone diagnostic measures classifier-free separability. JLT adds a head-only counterfactual intervention for decoder models: it freezes the incrementally trained backbone and asks how much performance can be recovered by re-aligning the LM head.

An interface is defined by what is frozen and which readout is used, not by whether the backbone weights are updated. The backbone diagnostic therefore applies whenever the effective representation \(\phi_t(x)\) produced by \(m_t\) changes across tasks. This includes prompt-based methods such as L2P~\citep{wang2022l2p}, whose prompts alter the hidden states, and LoRA-based methods such as InfLoRA~\citep{liang2024inflora}, whose low-rank updates are merged into the backbone. It becomes constant only when the forward pass never changes, as for SEQ*~\citep{seq*} after warm-up or KLDA~\citep{momeni2025continual}; the constant value itself indicates zero representation-level plasticity.

\input{tables/interface_estimands}

%% file: tables/interface_estimands.tex
\begin{table}[t]
\centering
\caption{Evaluation interfaces considered in this paper. Each interface instantiates the same BWT/FWT formulas but answers a different measurement question. Frozen denotes the component held fixed while a new readout is fitted at evaluation time; the trained head and the backbone diagnostic fit no new readout.}
\label{table:interface_estimands}

\small
\setlength{\tabcolsep}{4.5pt}
\begin{tabularx}{\textwidth}{@{}>{\raggedright\arraybackslash}p{0.115\textwidth}l>{\raggedright\arraybackslash\hsize=0.86\hsize}X>{\raggedright\arraybackslash\hsize=1.2\hsize}X>{\raggedright\arraybackslash\hsize=0.94\hsize}X@{}}
\toprule
\textbf{Interface} & \textbf{Frozen} & \textbf{Readout} & \textbf{Measures} & \textbf{Does not measure} \\
\midrule
Trained head & Nothing & Classifier or LM head & Deployed IL performance & Whether the backbone retained information \\
\addlinespace
Probe & Backbone & Newly trained classifier & Recoverable information under a fresh supervised readout & Deployed performance of the IL model \\
\addlinespace
Backbone diagnostic & Nothing & None (clustering, NCM, kNN) & Classifier-free proxy for representation separability & Task accuracy or calibrated prediction \\
\addlinespace
JLT & Backbone & LM head only & Recoverable performance under head re-alignment & Standard CIL performance \\
\bottomrule
\end{tabularx}
\end{table}

%% file: sections/3.Method.tex
\section{Diagnostic interfaces}
\label{sec:4}

\begin{figure}[t!]
    \centering
    \begin{subfigure}[b]{0.47\textwidth}
        \centering
        \includegraphics[width=\textwidth]{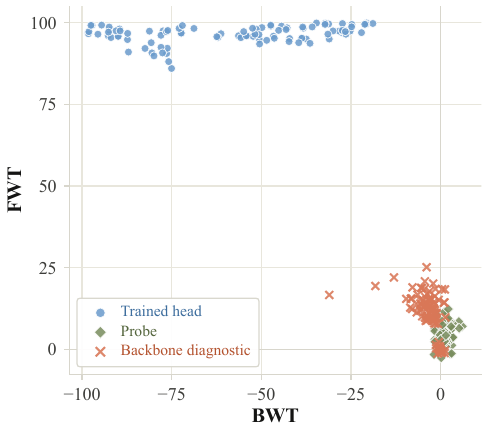}
        \caption{Discriminative backbone}
        \label{fig:encoder1}
    \end{subfigure}
    \begin{subfigure}[b]{0.47\textwidth}
        \centering
        \includegraphics[width=\textwidth]{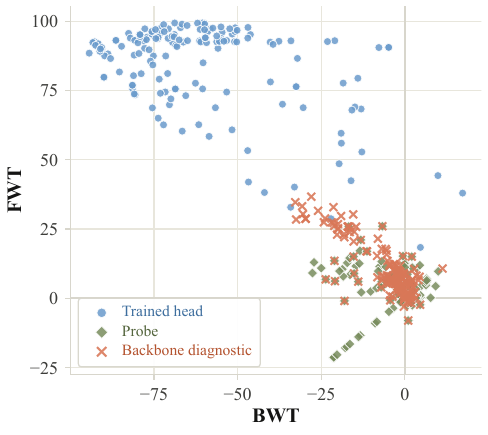}
        \caption{Generative backbone}
        \label{fig:decoder1}
    \end{subfigure}

    \caption{BWT and operational FWT evaluated under three interfaces. Results are aggregated over the benchmark streams in Table~\ref{table:dataset}; the number of incremental tasks varies by dataset.}
    \label{fig:encoder_decocer_figure1}
\end{figure}

\subsection{Head-level BWT and FWT}
\label{sec:4.1}
The most common evaluation interface is the trained head, the classifier or LM head that is updated sequentially with the backbone. This is the appropriate interface for reporting task accuracy, but its scores are difficult to attribute to a specific component. Prior work shows that sequentially trained classifiers can become strongly biased toward recently learned classes~\citep{hou2019learning, wu2019large, wu2022pretrained, seq*}. As a result, low accuracy on earlier tasks may reflect instability in the head, degradation in the backbone, or both.

Figure \ref{fig:encoder1} illustrates this ambiguity. BWT-head values differ substantially across IL methods, suggesting different degrees of forgetting. However, because the head itself is part of the sequentially updated system, these values do not determine whether the retained representation has disappeared or whether it is no longer read out correctly. Operational FWT-head is even less direct: before a new task is introduced, the head may have no meaningful output for unseen classes.


\subsection{Probe-level BWT and FWT}
\label{sec:4.2}
The probe is a linear classifier newly trained on frozen backbone features, as shown in Figure \ref{fig:architecture}. It removes the bias of the trained head and can indicate whether useful information remains accessible in the backbone \citep{hou2019learning, zhou2022forward, seq*}. However, probe-level evaluation answers a different question from standard task accuracy: it measures what a newly optimized classifier can recover from the representation, not what the incrementally trained model actually predicts.

In Figure \ref{fig:encoder1}, BWT-probe remains close to 0 for most models and IL methods. This suggests that much of the information needed for old tasks is still recoverable. At the same time, a strong probe may compensate for representation shifts or weak separability \citep{zhou2021directprobe}, thereby masking changes in the backbone itself.

For FWT, the backbone is frozen after each task and the probe is trained over the relevant classes. This makes it possible to evaluate future-task labels, but the measured score depends heavily on the newly trained probe. Thus, the probe measures recoverability under a new readout, but it cannot distinguish an unchanged representation from a shifted one that the probe can still decode.

\subsection{Backbone-level BWT and FWT}
\label{sec:4.3}
The backbone has no standalone prediction interface, so backbone-level transfer cannot be observed through accuracy in the same way as head-level transfer. We therefore define the backbone diagnostic as a classifier-free measure of representation separability, instantiated mainly by clustering. As shown in Figure \ref{fig:architecture}, we extract backbone representations after each task, cluster them, align clusters to labels, and use the resulting scores to instantiate BWT-backbone and FWT-backbone. This diagnostic is not intended to replace task accuracy. Instead, it approximates how separable task representations remain independently of any trained readout.

This choice follows a broader line of work that analyzes backbone features to reduce dependence on classifier bias \citep{liu2019continual, zhou2021closer, zhou2021directprobe}; related work also uses feature diversity as an auxiliary IL diagnostic \citep{chen2023forgetting}. Clustering is not a gold-standard measure of representation quality: it is sensitive to feature normalization, the number of clusters, class imbalance, cluster-label matching, the algorithm, and the metric. We therefore treat it as one diagnostic among several and stress-test it across algorithms, metrics, and settings (Section~\ref{sec:robust_clustering}).

We evaluate clustering diagnostics with five algorithms and three metrics: ACC, ARI, and NMI. Main visualizations use Spectral Clustering, while Table~\ref{table:interface_summary} aggregates all clustering algorithms and metrics. We complement clustering with diagnostics that involve no cluster count, initialization, label matching, or trained readout: nearest-class-mean (NCM) and \(k\)-nearest-neighbor accuracy, the silhouette score computed with ground-truth labels, and the Fisher discriminant ratio (Appendix~\ref{sec:additional_units}). Details are in Appendix~\ref{appendix:B}.




%% file: sections/4.Experimental_Setup.tex
\section{Experimental setup}
\label{sec:experimental_setup}
\subsection{Task \& baseline}
We evaluate five benchmark groups spanning seven sentence-level datasets, as shown in Table \ref{table:dataset}. Each benchmark group defines its own incremental stream, with task counts ranging from 5 to 15.
For text classification, we use three datasets, AGNews, DBPedia, and Yahoo Answers, as Topic3Datasets~\citep{yahoo}. For intent classification, we use CLINC150~\citep{Clinc150} and Banking77~\citep{banking77}. Finally, for relation extraction, we use FewRel~\citep{fewrel} and TACRED~\citep{tacred}.


For discriminative backbones, we adopt four representative IL methods: Base, ER~\citep{er}, DER++~\citep{derpp}, and CLSER~\citep{clser}. For generative backbones, we compare eight baselines: Base, DER++, CLSER, L2KD~\citep{l2kd}, LAMOL\_g, LAMOL\_t~\citep{lamol}, LAMOL\_KD~\citep{seq*}, and PCLL~\citep{pcll}. We also report recent competitive frozen-backbone methods, such as SEQ* \citep{seq*} and KLDA \citep{momeni2025continual}, in the JLT performance table (Appendix Tables~\ref{tab:table_main} and~\ref{tab:table_main_b}). Because these methods keep the backbone fixed after warm-up or use fixed embeddings, their BWT-backbone and FWT-backbone are constant by construction, so they are not included in the backbone-level correlation analysis. For detailed descriptions of the baselines, please refer to Appendix \ref{appendix:C}.

\subsection{Backbone}
We evaluate both discriminative backbones (encoder architectures) and generative backbones (decoder architectures). For discriminative backbones, we use BERT-base, BERT-large~\citep{bert}, RoBERTa-base, and RoBERTa-large~\citep{roberta}. For generative backbones, we use Pythia~\citep{pythia} models based on GPT-NeoX~\citep{gptneox} in different sizes (70M, 160M, 410M), as well as Qwen2-0.5B~\citep{yang2024qwen2technicalreport}, Qwen2.5-0.5B~\citep{qwen2025qwen25technicalreport}, and Qwen3-0.6B~\citep{yang2025qwen3}. For model-type and model-size analysis, we additionally evaluate the Base method on Pythia-1.4B/2.8B/6.9B, Llama-3.2-1B/3B, Llama-3.1-8B~\citep{dubey2024llama}, and Qwen3-0.6B/1.7B/4B/8B. For clustering-free diagnostics and parameter-efficient methods, we further evaluate 27 backbone \(\times\) dataset \(\times\) method units on Llama-3.2-1B and Qwen3-0.6B, including LoRA~\citep{lora} and prompt-tuning variants (Appendix~\ref{sec:additional_units}). More details on training and backbone evaluation are in Appendix \ref{Appendix_A.1}.

%% file: sections/5.Results.tex
\section{Results}

\input{tables/interface_summary}

\subsection{The same BWT/FWT metric yields different conclusions}
\label{sec:5.1}
The three interfaces do not merely differ in magnitude; they support different qualitative conclusions. In discriminative backbones, the per-method plots in Appendix Figure~\ref{fig:encoder3} show that BWT-head suggests severe forgetting, especially for Base, because the trained head is strongly biased toward recently learned classes. BWT-probe remains close to 0, suggesting that a newly trained probe can recover much of the old-task information. BWT-backbone lies between these two views: the representation itself changes, but not as drastically as head-level evaluation implies.

The operational FWT results show the same interface dependence. FWT-head values can exceed 80, largely reflecting the introduction of task-specific output classes rather than pure transfer. FWT-probe remains small (below 5 on average) because the probe can compensate for representation differences. By contrast, FWT-backbone ranges over a meaningful intermediate scale, indicating measurable changes in representation separability before and after task training.

\begin{figure}[t!]
    \centering
    \begin{subfigure}[b]{0.32\textwidth}
        \centering
        \includegraphics[width=\textwidth]{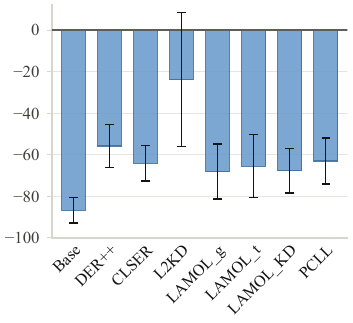}
        \caption{BWT-head}
        \label{fig:Decoder_BWT}
    \end{subfigure}
    \begin{subfigure}[b]{0.32\textwidth}
        \centering
        \includegraphics[width=\textwidth]{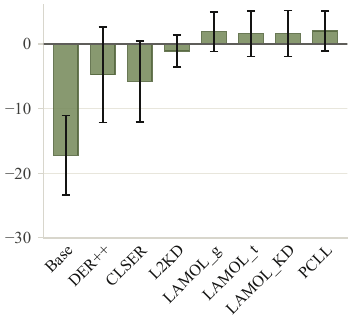}
        \caption{BWT-probe}
        \label{fig:Decoder_BWT-Prob}
    \end{subfigure}
    \begin{subfigure}[b]{0.32\textwidth}
        \centering
        \includegraphics[width=\textwidth]{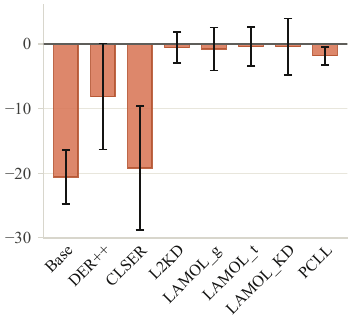}
        \caption{BWT-backbone}
        \label{fig:Decoder_BWT-Clustering}
    \end{subfigure}
    \begin{subfigure}[b]{0.32\textwidth}
        \centering
        \includegraphics[width=\textwidth]{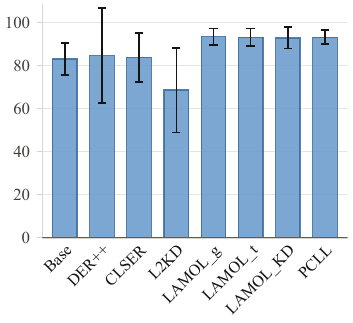}
        \caption{FWT-head}
        \label{fig:Decoder_FWT}
    \end{subfigure}
    \begin{subfigure}[b]{0.32\textwidth}
        \centering
        \includegraphics[width=\textwidth]{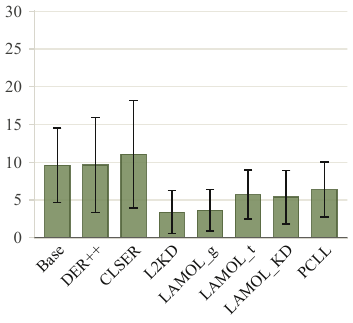}
        \caption{FWT-probe}
        \label{fig:Decoder_FWT-Prob}
    \end{subfigure}
    \begin{subfigure}[b]{0.32\textwidth}
        \centering
        \includegraphics[width=\textwidth]{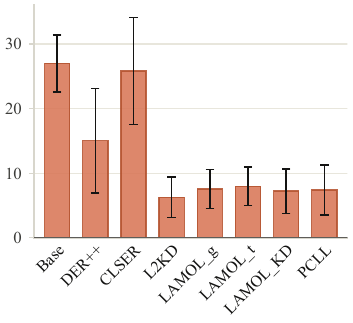}
        \caption{FWT-backbone}
        \label{fig:Decoder_FWT-Clustering}
    \end{subfigure}
    \caption{Average and standard deviation of head-level, probe-level, and backbone-level BWT/FWT for each IL method in six generative backbones, aggregated over the benchmark streams in Table~\ref{table:dataset}.}
    \label{fig:decoder3}
\end{figure}

The generative-backbone results in Figures~\ref{fig:decoder1} and~\ref{fig:decoder3} mirror this pattern. BWT-head again indicates severe forgetting (e.g., $-86.7$ on average for Base), while the backbone diagnostic shows that decoder representations retain substantially more structure than the LM head can read out directly. This suggests that the backbone--head gap is not limited to encoder classifiers.

\paragraph{Finding 1.}
The trained head suggests catastrophic forgetting, the probe suggests strong recoverability, and the backbone diagnostic indicates partial representation drift. Forgetting is therefore not a single model-level number; it depends on the interface through which it is measured.

\subsection{Backbone trends are consistent across diagnostic choices}
\label{sec:robust_clustering}
BWT and FWT share the intermediate term \(a_{i,i}\):
\[
BWT_i = a_{T,i} - a_{i,i}, \qquad
FWT_i = a_{i,i} - a_{i-1,i}.
\]
This coupling can make a negative correlation partly expected by construction, so the sign alone is not informative. What matters is how consistently and how strongly the relation holds across interfaces. As Table~\ref{table:interface_summary} shows, the trained head and probe interfaces have mixed signs and weak aggregate correlations, whereas the backbone diagnostic is negative in every clustering algorithm, metric, and backbone combination (mean \(r=-0.74\)). The 150 configurations are related rather than independent samples, so we use them to assess robustness, not to estimate a sampling distribution.

Because clustering scores depend on normalization, the number of clusters, initialization, and label matching, we also evaluate clustering-free diagnostics on 27 additional backbone \(\times\) dataset \(\times\) method units (Appendix Table~\ref{tab:additional_units}). NCM stays strongly negative in every subset (\(-0.56\) to \(-0.92\)), whereas the trained-head correlation changes sign (\(-1.00\) to \(+0.33\)). Over all 27 units, every clustering-free diagnostic is negative (NCM \(-0.87\), silhouette \(-0.84\), kNN-10 \(-0.80\), kNN-1 \(-0.62\), Fisher ratio \(-0.50\)), and the clustering-based correlation remains negative under all 25 settings of normalization, cluster count, seed, label matching, class balance, sample size, and algorithm, for ACC, ARI, and NMI alike (Appendix~\ref{sec:sensitivity}). Clustering-based and clustering-free diagnostics therefore agree, and the representation-level relation does not hinge on clustering. Full Pearson and Spearman correlations for the main study are reported in Tables~\ref{table:pearson} and~\ref{table:spearman}.
\paragraph{Finding 2.}
The representation-level stability--plasticity pattern is consistent across diagnostic choices, while the trained-head and probe interfaces remain unstable in sign and magnitude.

\subsection{Stability--plasticity trade-off in the backbone}
\label{sec:sp_backbone}

Strong IL results can suggest that the stability--plasticity dilemma has largely been addressed, but many of these gains come from keeping the backbone close to its earlier state while adjusting the classifier or LM head. Our backbone diagnostic suggests that the trade-off persists at the representation level: BWT-backbone and FWT-backbone are negatively correlated across backbones and diagnostics (Table~\ref{table:interface_summary}), so configurations that retain more old-task structure tend to gain less plasticity on new tasks. Better head-level performance therefore does not by itself show that the backbone keeps learning new tasks effectively.

\paragraph{Parameter-efficient methods.}
The diagnostic also applies to parameter-efficient fine-tuning (PEFT), because LoRA updates and prompts change the effective representation \(\phi_t(x)\) (Section~\ref{sec:estimands}). On Llama-3.2-1B, LoRA still shows severe forgetting at the trained head (BWT-head \(-76.5\) to \(-89.0\)) and prompt tuning keeps head accuracy low throughout (\(A_T\) 4.6--10.2), while the representation-level changes are much smaller (Appendix Table~\ref{tab:peft_units}). On Banking77, LoRA is more stable than full fine-tuning with Base at the backbone (BWT-backbone \(-1.91\) vs.\ \(-3.71\)) but has less than half of its plasticity (FWT-backbone \(2.70\) vs.\ \(6.06\), and \(16.89\) for CLSER), although all three methods show BWT-head below \(-82\). Parameter isolation thus buys stability at the cost of representation-level plasticity, a trade-off that is invisible at the trained head.

\paragraph{Finding 3.} The backbone diagnostic does not support the view that current IL methods have resolved the stability--plasticity dilemma at the representation level.




\subsection{Secondary observation: scaling improves retention more than plasticity}
\label{sec:scaling_observation}
In the Base setting, increasing model size within Pythia, Llama, and Qwen improves BWT-backbone more consistently than FWT-backbone (Appendix Figure~\ref{fig:capacity}), suggesting that scale helps retention more than the immediate plasticity gain measured by our operational FWT.

%% file: tables/interface_summary.tex
\begin{table}[t]
\centering
\caption{BWT--FWT Pearson correlation at each interface in the CIL scenario. The 150 configurations of the backbone diagnostic combine 10 backbones, five clustering algorithms, and three metrics; they are related rather than independent, so Std summarizes their spread rather than a standard error. Full Pearson and Spearman results are in Appendix~\ref{appendix:E}.}
\label{table:interface_summary}
\small
\setlength{\tabcolsep}{6pt}
\begin{tabular}{@{}llcccl@{}}
\toprule
\textbf{Interface} & \textbf{Configurations} & \textbf{Mean \(r\)} & \textbf{Std} & \textbf{Range} & \textbf{Sign pattern} \\
\midrule
Trained head & 10 backbones & $-$0.13 & 0.38 & $-$0.52 to 0.43 & Mixed \\
Probe & 10 backbones & $-$0.19 & 0.44 & $-$0.62 to 0.65 & Mixed \\
Backbone diagnostic & 150 (10 \(\times\) 5 \(\times\) 3) & $-$0.74 & 0.13 & $-$0.93 to $-$0.43 & Negative in all 150 \\
\bottomrule
\end{tabular}
\end{table}

%% file: sections/6.Analysis.tex
\section{A Head-Only Intervention Estimates Recoverable Forgetting}
\label{sec:jlt}

The three-interface analysis suggests that poor standard accuracy may not imply that the backbone has erased the task. To test this interpretation, after an IL method has trained the model sequentially, we freeze the backbone and re-optimize only the LM head with a label-token cross-entropy objective, without adding adapters, external encoders, or independent classifiers (Appendix~\ref{sec:lmhead_details}; full results in Appendix Tables~\ref{tab:table_main}--\ref{tab:table_main_b}).

\input{tables/jlt_gap_summary}

JLT is not meant to compete with standard IL protocols but to measure how much performance would be recoverable if only the readout were re-aligned. We therefore report the JLT gap,
\[
G_{\mathrm{JLT}} = A_T^{\mathrm{JLT}} - A_T^{\mathrm{standard}},
\]
as the primary quantity. A large gap indicates that the loss in standard accuracy is not fully explained by representation erasure; a small gap combined with low standard accuracy would instead indicate that the backbone itself has lost the relevant information.

The aggregate results in Table~\ref{table:jlt_gap_summary} show that all four IL methods have large recoverable gaps. Although standard evaluation indicates severe forgetting, freezing the backbone and re-aligning only the LM head recovers much of the performance. For example, L2KD averages \(40.72\) under standard evaluation but \(95.56\) under the diagnostic head-realignment protocol, reducing average BWT from \(-42.75\) to \(-1.76\). Over the same models and benchmarks, joint fine-tuning averages \(96.38\) in \(A_T\), while L2KD+JLT averages \(95.56\), motivating a diagnostic decomposition of the gap to the joint-training reference.

Let \(A_T^{\mathrm{joint}}\) denote the joint-training reference. The gap from standard IL evaluation to the joint reference can be decomposed as
\[
A_T^{\mathrm{joint}}-A_T^{\mathrm{standard}}
=
\underbrace{A_T^{\mathrm{JLT}}-A_T^{\mathrm{standard}}}_{\text{readout-recoverable gap}}
+
\underbrace{A_T^{\mathrm{joint}}-A_T^{\mathrm{JLT}}}_{\text{residual gap}}.
\]
We use this decomposition only diagnostically. The first term estimates the portion of apparent forgetting that can be removed by re-aligning the LM head on a fixed incrementally trained backbone; the second term is the residual gap that head re-alignment alone does not remove. This decomposition shows that recoverability is method-dependent (Table~\ref{table:jlt_gap_summary}). L2KD, LAMOL\_KD, and PCLL recover 96--99\% of the gap to the joint reference, whereas Base recovers only 74.6\% (74.0--75.2\% across the three backbones). The residual is also stream-specific: for Base it is close to zero on TACRED (\(-0.3\) to \(1.2\)) but about 44 points on FewRel for every backbone, while for L2KD it is at most 4.34 points on every stream (Appendix Table~\ref{tab:jlt_residual}). Neither pattern is visible from standard accuracy or from \(G_{\mathrm{JLT}}\) alone.

\input{tables/jlt_taxonomy}

Table~\ref{tab:jlt_taxonomy} turns the two terms into a two-axis reading. A method with poor standard accuracy but a large \(G_{\mathrm{JLT}}\) and a small residual, such as L2KD, is readout-limited: its representation remains usable, and improving it requires changes to the head rather than to the backbone. A large residual instead calls for backbone-level intervention, as for Base on FewRel.

\paragraph{Effect of head re-alignment on FWT.}
Under task-adaptive JLT, both \(a_{i,i}\) and \(a_{i-1,i}\) are measured with re-aligned heads, so JLT-FWT excludes the gain that comes from creating an output interface for previously unseen labels. On the relation-extraction streams (TACRED and FewRel), FWT drops by 29.5--40.4 points on average for every method; for Pythia-410M with L2KD on TACRED, it falls from 94.39 to 4.92. On the intent and topic streams, the change is small for Base, LAMOL\_KD, and PCLL (\(-4.3\) to \(-1.1\)), and L2KD even gains 21.5 points, mainly on Topic3, where its standard FWT is only about 40 (Appendix Table~\ref{tab:jlt_residual}). Most of the head-level FWT on relation extraction therefore reflects interface construction, whereas on intent and topic streams it largely reflects plasticity that persists after the readout is aligned.

These results should not be read as evidence that head-only tuning is a deployable replacement for continual learning. The task-adaptive JLT variant adapts the head before evaluating each task and therefore estimates recoverable performance rather than standard CIL accuracy. A deployable variant, post-hoc JLT, re-aligns the head once per checkpoint under the same rehearsal assumption as ER, using a small buffer drawn from the tasks seen so far with roughly one example per class, and never accesses the full training split of an evaluated task. On Llama-3.2-1B, it improves standard \(A_T\) by 23.8 points and BWT by 52.6 points on average over four dataset--method pairs, while FWT falls from 94.3 to 70.7 as the interface-construction component is removed (Appendix~\ref{sec:posthoc_jlt}).

\paragraph{Finding 4.}
With the incrementally trained backbone fixed, re-aligning only the LM head recovers 96--99\% of the gap to joint training for replay- and distillation-based methods but only about three quarters for Base, whose unrecovered residual concentrates on specific streams such as FewRel.

\paragraph{Threats to interpretation.}
Our backbone diagnostic is not task accuracy; it is a classifier-free separability proxy. The clustering configurations are not independent statistical samples, but robustness stress tests over related diagnostic choices. Because BWT and operational FWT share \(a_{i,i}\), we interpret their relation comparatively across interfaces rather than as proof of a universal trade-off. Finally, task-adaptive JLT is oracle-style: it measures recoverability under head re-alignment, not deployable CIL performance.

%% file: tables/jlt_gap_summary.tex
\begin{table}[!htb]
\centering
\caption{Aggregate JLT decomposition over three decoder backbones and five benchmark groups (task-adaptive protocol; Appendix Tables~\ref{tab:table_main}--\ref{tab:table_main_b}). \mbox{\(G_{\mathrm{JLT}} = A_T^{\mathrm{JLT}} - A_T^{\mathrm{standard}}\)}, \mbox{Residual \(= A_T^{\mathrm{joint}} - A_T^{\mathrm{JLT}}\)} with \mbox{\(A_T^{\mathrm{joint}} = 96.38\)}, and \mbox{Recovery \(= G_{\mathrm{JLT}} / (A_T^{\mathrm{joint}} - A_T^{\mathrm{standard}})\)}.}
\label{table:jlt_gap_summary}

\small
\setlength{\tabcolsep}{4pt}
\begin{tabular}{@{}lccccccccc@{}}
\toprule
 & \multicolumn{5}{c}{\textbf{\boldmath\(A_T\)}} & \multicolumn{2}{c}{\textbf{BWT}} & \multicolumn{2}{c}{\textbf{FWT}} \\
\cmidrule(lr){2-6}\cmidrule(lr){7-8}\cmidrule(l){9-10}
\textbf{IL method} & Standard & JLT & \(G_{\mathrm{JLT}}\) & Residual & Recovery & Standard & JLT & Standard & JLT \\
\midrule
Base & 11.07 & 74.73 & +63.65 & 21.65 & 74.6\% & $-$87.79 & $-$19.12 & 86.55 & 70.91 \\
L2KD & 40.72 & 95.56 & +54.84 & 0.82 & 98.5\% & $-$42.75 & $-$1.76 & 74.77 & 75.86 \\
LAMOL\_KD & 40.63 & 94.14 & +53.50 & 2.25 & 96.0\% & $-$62.39 & $-$2.68 & 93.19 & 74.47 \\
PCLL & 41.63 & 94.51 & +52.88 & 1.87 & 96.6\% & $-$61.48 & $-$1.89 & 90.48 & 76.21 \\
\bottomrule
\end{tabular}
\end{table}

%% file: tables/jlt_taxonomy.tex
\begin{table}[t]
\centering
\caption{Two-axis reading of the JLT decomposition. Examples use Table~\ref{table:jlt_gap_summary} and Appendix Table~\ref{tab:jlt_residual}.}
\label{tab:jlt_taxonomy}

\small
\renewcommand{\arraystretch}{1.1}
\begin{tabularx}{\textwidth}{@{}lYY@{}}
\toprule
 & \textbf{Small residual} (representation preserved) & \textbf{Large residual} (representation lost) \\
\midrule
\textbf{Large \(G_{\mathrm{JLT}}\)} & \textit{Readout-limited}: realign or preserve the head (e.g., L2KD, residual 0.82) & \textit{Mixed}: intervene on both head and backbone (e.g., Base, residual 21.65; FewRel \(\approx\)44) \\
\addlinespace
\textbf{Small \(G_{\mathrm{JLT}}\)} & \textit{Already aligned} & \textit{Representation-limited}: replay, regularization, or another backbone-level intervention \\
\bottomrule
\end{tabularx}
\end{table}

%% file: sections/7.Conclusion.tex
\section{Conclusion}
Our central message is that forgetting in incremental learning is not a single model-level quantity. It is an interface-dependent measurement. The trained head measures deployed behavior, the probe measures recoverability under a new readout, and the backbone diagnostic measures classifier-free separability. These views can disagree sharply, and treating them as interchangeable can lead to misleading conclusions about stability and plasticity. The JLT intervention shows why this distinction matters: when the incrementally trained backbone is fixed and only the LM head is re-aligned, much of the apparent forgetting is recoverable. This does not mean that standard IL is solved, nor that head-only tuning is a deployable replacement for continual learning. It means that many failures attributed to catastrophic forgetting are better understood as a combination of partial representation drift and readout mismatch. We therefore recommend that future incremental learning work report not only BWT, FWT, and average accuracy, but also the interface through which these quantities are measured.

%% file: sections/8.Appendix.tex
\renewcommand{\thesection}{\Alph{section}}

\section*{Limitations}
We use classifier-free diagnostics to approximate changes in backbone separability, but they are not a complete substitute for head-level task accuracy. Clustering scores in particular depend on the algorithm, number of clusters, feature normalization, class balance, cluster-label matching, and evaluation metric, which is why we also report clustering-free diagnostics and a sensitivity analysis. Our goal is therefore not to propose a perfect new evaluation metric, but to provide an auxiliary view of how the backbone representation changes during IL.

JLT also has an important limitation: the task-adaptive variant performs head-only adaptation before evaluating each task. This should be interpreted as an oracle-style diagnostic of recoverable information in the backbone, not as a deployable CIL protocol that has access to each test task's training data at evaluation time. The reported gains are useful for quantifying head--backbone mismatch, but they should not be read as direct standard IL performance.

Sequential fine-tuning across many tasks is computationally expensive, which limited our experiments with larger generative models. We attempted to use Qwen3-32B \citep{yang2025qwen3}, Llama-3-70B \citep{dubey2024llama}, and related models, but could not run them within our compute budget or apply every IL method under the same conditions. Models larger than 8B typically require PEFT \citep{houlsby2019parameter, lora, han2024parameter}; in that setting, several IL methods (L2KD, LAMOL, LAMOL\_KD) did not work reliably. Our experiments are therefore limited to models below 10B parameters. Proposing a new IL method that optimizes the backbone-level stability--plasticity trade-off remains future work.

\input{sections/impact_statement}

\section{Experimental details}
\label{Appendix_A}



\subsection{Backbone details}
\label{Appendix_A.1}
For the discriminative backbone, learning and evaluation are performed using the [CLS] token features of the last hidden states \citep{ethayarajh2019contextual}. The classifier for each task is a linear layer whose input dimension is the hidden size of the backbone and whose output dimension is the number of classes in that task. In TIL, each task is trained and evaluated with its own classifier; in CIL, the logits of all task classifiers are concatenated. To use the generative backbone in sentence-level tasks, we use two types of prompts depending on the task type. For text and intent classification, we use the following prompt:

\noindent
\texttt{"Input sentence: \{text\}\textbackslash n The Label: \{label\}\{eos token\}"}

For relation extraction, we use the following prompt:

\noindent
\texttt{"Input sentence: \{text\}\textbackslash n The relationship between \{head entity\} and \{tail entity\} is \{label\}\{eos token\}"} 

With these prompts, we optimize the causal language modeling loss on the \texttt{\{label\}\{eos token\}} span \citep{seq*}. Depending on the IL method, additional objectives such as cross-entropy, KL-divergence, or MSE distillation losses are used.




\subsection{Implementation details}
\label{Appendix_A.2}
For Topic3Datasets, we use a maximum sequence length of 256 and 3 epochs per incremental task; for FewRel and TACRED, 128 and 5 epochs; and for CLINC150 and Banking77, 64 and 5 epochs. We used a learning rate of $1 \times 10^{-5}$ for each backbone and $1 \times 10^{-3}$ for the classifier with the AdamW optimizer \citep{DBLP:journals/corr/abs-1711-05101}. Probes are linear classifiers trained for 20 epochs on frozen backbone features.

All experiments were run on eight NVIDIA RTX A5000 GPUs (24GB each). Unless otherwise noted, each experiment was run with three random seeds, reported means average over the seeds, and error bars in figures denote one standard deviation across seeds. Correlation coefficients are computed over the corresponding dataset--method units for each evaluation interface. The compute estimates in Table~\ref{tab:compute_resources} are approximate and are intended to support reproducibility planning rather than exact accounting. For the remaining hyperparameters of each IL method, we follow the settings recommended in the original papers.

\input{tables/compute_resources}
\input{tables/dataset}
For assets whose public dataset cards do not specify a single explicit license, we use the datasets only under their commonly distributed research benchmark setting and cite the original sources. We do not redistribute the raw datasets.

\input{tables/assets_license}

\subsection{Evaluation metrics}
\subsubsection{Average accuracy}
\label{Average_Accuracy}
The most common way to evaluate IL is to measure performance on all tasks after the final task has been learned, and Average Accuracy is reported in most studies. When $T$ tasks are evaluated, however, $T-1$ of them are previous tasks, so this metric rewards retention much more than learning the current task. Average Accuracy after task $t$ is defined as
\begin{equation}
\label{equation:Acc}
A_t = \frac{1}{t} \sum_{i=1}^{t} a_{t,i}
\end{equation}
where $a_{t,i}$ is the accuracy on the test set of task $i$ after the model has been trained incrementally on tasks $1$ to $t$.

\subsubsection{Average incremental accuracy}
\label{Average_Incremental_Accuracy}
Average Incremental Accuracy averages $A_t$ over all steps $t=1,\dots,T$. It avoids judging a method by $A_T$ alone, but it usually follows the same trend as Average Accuracy because previous-task accuracy still dominates the average.
\begin{equation}
\bar{A} = \frac{1}{T} \sum_{t=1}^{T} A_t
\end{equation}

\section{Clustering algorithm details}
\label{appendix:B}
To evaluate the features extracted from the backbone, we employ five representative clustering algorithms: K-means~\citep{kmeans}, Gaussian Mixture Model~\citep{GaussianMixtureModel}, Spectral Clustering~\citep{spectral}, Agglomerative Clustering~\citep{agglomerative}, and Deep Clustering~\citep{deepcluster}. While all algorithms are evaluated, the main visualizations use Spectral Clustering and the aggregate summary in Table~\ref{table:interface_summary} uses all algorithms and metrics.

\subsection{Clustering evaluation metrics}
\label{appendix:B_1}

\noindent
\textbf{ACC (Accuracy)}

We build a contingency matrix between predicted clusters and ground-truth labels, find the one-to-one cluster-to-label mapping that maximizes agreement with the Hungarian algorithm, and report accuracy under this mapping.

\noindent
\textbf{ARI (Adjusted Rand Index)}

ARI \citep{hubert1985comparing} measures the agreement between a clustering and the ground-truth labels over pairs of data points, corrected for chance. It equals 1 for perfect agreement and 0 in expectation for a random clustering; negative values indicate agreement below chance \citep{vinh2009information, zhang2019critical}. With the pair counts defined below, ARI can be written as

\begin{equation}
ARI=\frac{2 \cdot (TP \cdot TN - FN \cdot FP)}{(TP + FN) \cdot (FN + TN) + (TP + FP) \cdot (FP + TN)}
\end{equation}

\noindent
\textbf{TP (true positive)}: the number of sample pairs placed in the same cluster that also share the same ground-truth label.

\noindent
\textbf{TN (true negative)}: the number of sample pairs placed in different clusters that also have different labels.

\noindent
\textbf{FP (false positive)}: the number of sample pairs placed in the same cluster but with different labels.

\noindent
\textbf{FN (false negative)}: the number of sample pairs placed in different clusters but with the same label.

\noindent
\textbf{NMI (Normalized Mutual Information)}

NMI measures the mutual information between the cluster assignments and the ground-truth labels, normalized by their entropies. It ranges from 0 to 1, with values closer to 1 indicating stronger dependence between clusters and labels.

\begin{equation}
I(U, V) = \sum_{u \in U} \sum_{v \in V} P(u, v) \cdot \log \left( \frac{P(u, v)}{P(u) \cdot P(v)} \right)
\end{equation}
\begin{equation}
H(U) = - \sum_{u \in U} P(u) \cdot \log P(u)
\end{equation}

\begin{equation}
NMI(U, V) = \frac{I(U, V)}{\sqrt{H(U)H(V)}}
\end{equation}

Here $U$ denotes the cluster assignment and $V$ the ground-truth labels, $I(U,V)$ is their mutual information, and $H(\cdot)$ is entropy ($H(V)$ is defined analogously to $H(U)$). NMI normalizes $I(U,V)$ by the geometric mean of $H(U)$ and $H(V)$.

\input{tables/roberta_base_er}

\subsection{Experiment results by clustering algorithm}
We evaluated all models and IL methods with five clustering algorithms and three metrics. Most clustering algorithms show broadly similar patterns. For comparison with the commonly reported Average Accuracy of the trained head, we present the results of experiments with ER method on RoBERTa-base model as a representative of discriminative backbone in Table \ref{table3}. Similarly, we present the results of experiments with LAMOL\_g method on Pythia-410M model as a representative of generative backbone in Table \ref{table5}. $A_{T}$ is the trained head's Average Accuracy over all tasks after the final task $T$ (Appendix~\ref{Average_Accuracy}).

In the CIL scenario, the clustering scores are often higher than the trained head's $A_T$. For RoBERTa-base with ER (Table~\ref{table3}), clustering ACC exceeds $A_T$ for all five algorithms on the four datasets with seven or more incremental tasks (TACRED, Banking77, CLINC150, and FewRel), whereas on Topic3Datasets, which has only five tasks, $A_T$ is higher than the ACC of every algorithm. For Pythia-410M with LAMOL\_g (Table~\ref{table5}), clustering ACC exceeds $A_T$ on Banking77, CLINC150, and FewRel but is close to or slightly below $A_T$ on TACRED. This pattern suggests that as the number of incremental tasks increases, head-level average accuracy can become increasingly affected by forgetting in the head. The clustering results should not be interpreted as higher task accuracy, but as evidence that representation separability can remain higher than what the trained head exposes.

Our goal in this study is not to determine which clustering algorithm is best or which metric is the best. We therefore use Spectral Clustering, whose scores lie in the middle of the range, for the visualizations and analysis in the main paper; all results are reported in Appendix~\ref{appendix:E}.
\input{tables/pythia_410m_lamol_g}

\subsection{Sensitivity of the clustering diagnostic}
\label{sec:sensitivity}
Using the features stored for the 27 additional units, we recompute the clustering diagnostic under 25 settings. Starting from the default (Spectral, no normalization, \(k=C\), seed 0, Hungarian matching, natural class distribution, \(n=1{,}500\)), each setting changes one factor: normalization (L2, standardization, whitening), number of clusters (\(0.5C\), \(2C\)), seed (1, 2), label matching (majority vote), class balance (balanced subsampling), sample size (750, 3{,}000, full test set), and algorithm (K-means, GMM, Agglomerative); we also cross the three alternative algorithms with the four normalizations. Because all settings use identical stored features, differences reflect diagnostic choices rather than training variation. Table~\ref{tab:sensitivity} shows that the BWT--FWT correlation is negative in every setting for ACC, ARI, and NMI.

\input{tables/sensitivity}

\section{Incremental learning methods}
\label{appendix:C}
We measured BWT and FWT for representative Incremental Learning methods. Aside from a brief explanation, we adopted detailed experimental settings widely used in prior studies.

\textbf{Base} - Sequentially fine-tunes the model on each task without any mechanism against forgetting. When evaluated through the trained head, it tends to predict mostly the classes of the most recently trained task.

\textbf{ER}~\citep{er} - A classical replay technique that incorporates old samples to improve retention. When learning a new task, a portion of old samples is included in the training process. In this study, we include one old sample per class during training.

\textbf{DER++}~\citep{derpp} - DER++ extends ER by also storing the model's logits for buffered samples and matching them with an MSE loss, in addition to replaying their labels. Although originally developed for the computer vision domain, we evaluated this method with both discriminative and generative backbones.

\textbf{CLSER}~\citep{clser} - CLS-ER extends experience replay with two semantic memories, a plastic and a stable model maintained as exponential moving averages of the working model, and regularizes the working model toward them. Like DER++, this method was initially used in the computer vision domain, but we applied it to both discriminative and generative backbones.

\textbf{L2KD}~\citep{l2kd} - L2KD builds on LAMOL and adds knowledge distillation: for each new task, a teacher model is first trained on that task and then distilled into the lifelong learner.

\textbf{LAMOL\_g \& LAMOL\_t}~\citep{lamol} - LAMOL is a method designed for generative models. When training on a new task, the model generates pseudo-samples of previous tasks and learns them together with the new task. The two variants differ in the generation token: LAMOL\_g uses a single shared generation token for all tasks, whereas LAMOL\_t uses a task-specific token.

\textbf{LAMOL\_KD}~\citep{seq*} - Similar to L2KD, except that the teacher model is trained on all previous tasks and transfers this knowledge through knowledge distillation.

\textbf{PCLL}~\citep{pcll} - PCLL builds on LAMOL-style generative replay, using a prompt-conditioned variational autoencoder (VAE) to generate pseudo samples of previous tasks, combined with knowledge distillation.

\textbf{SEQ*}~\citep{seq*} - SEQ* reduces backbone forgetting and classifier bias by freezing the backbone after warming up on the first task and training only the classifier for subsequent tasks using frozen backbone outputs. Since the backbone remains fixed after the first task, its representation-level diagnostic scores do not change. This makes SEQ* useful as a frozen-backbone baseline, but its BWT-backbone and FWT-backbone are always 0 by construction.

\textbf{KLDA}~\citep{momeni2025continual} - KLDA projects input data into a high-dimensional feature space using a kernel function and then performs Linear Discriminant Analysis (LDA) to maximize class separability. It operates on fixed embeddings extracted from a foundation model, so the backbone is not updated during continual learning. We report its ensemble variant, KLDA-E.

\section{Results in the TIL scenario}
\label{appendix:D}
\begin{figure}[t]
    \centering
    \includegraphics[width=0.68\linewidth]{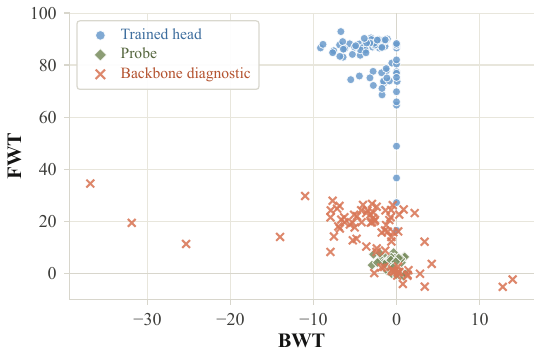}
    \caption{TIL sanity check. The x-axis represents BWT and the y-axis represents operational FWT; colors denote the three evaluation interfaces. Even when task identity restricts the prediction space, the interfaces remain separated, supporting that the backbone--head gap is not only CIL classifier bias.}
    \label{fig:til_sanity}
\end{figure}

The interface discrepancy in the CIL results could in principle arise only from class-incremental classifier bias. The TIL setting tests this explanation, because task identity restricts the prediction space. Figure~\ref{fig:til_sanity} shows that the interface discrepancy remains: head-level and probe-level evaluations become less dominated by global class competition, but the backbone diagnostic still shows the most consistent stability--plasticity pattern among the evaluated interfaces. The backbone--head gap is therefore broader than class-incremental classifier bias.

\begin{figure}[t!]
    \centerline{\includegraphics[width=\columnwidth]{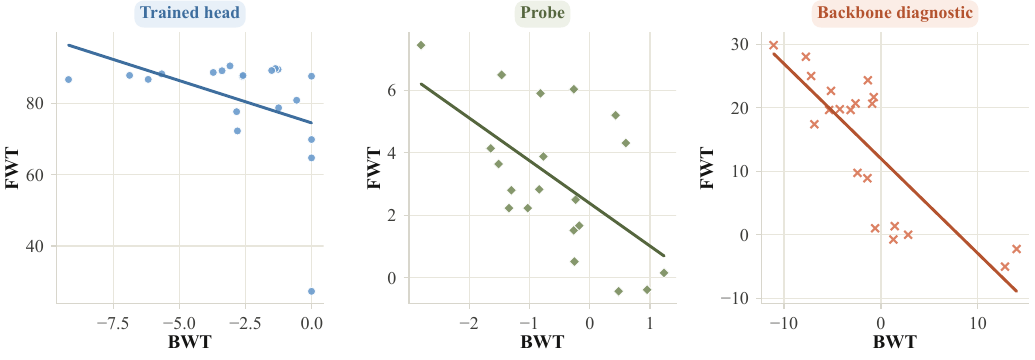}}
    \caption{BWT versus operational FWT under the three interfaces for RoBERTa-base in the TIL scenario (Base, ER, DER++, and CLSER), with least-squares fits. Correlation coefficients are reported in Table~\ref{table:til}.}
    \label{fig:til1} 
\end{figure}

\begin{figure}[!t]
    \centering
    \begin{subfigure}[b]{0.32\textwidth}
        \centering
        \includegraphics[width=\textwidth]{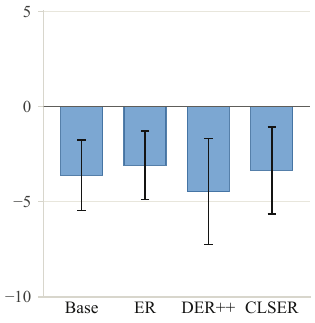}
        \caption{BWT-head}
        \label{fig:Encoder_BWT_appendix}
    \end{subfigure}    
    \begin{subfigure}[b]{0.32\textwidth}
        \centering
        \includegraphics[width=\textwidth]{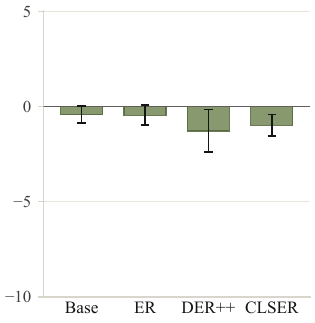}
        \caption{BWT-probe}
        \label{fig:Encoder_BWT-Prob_appendix}
    \end{subfigure}
    \begin{subfigure}[b]{0.32\textwidth}
        \centering
        \includegraphics[width=\textwidth]{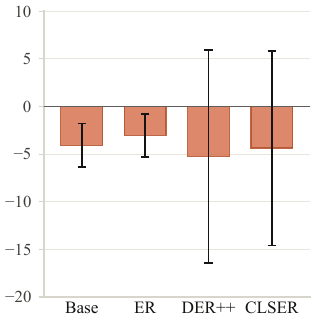}
        \caption{BWT-backbone}
        \label{fig:Encoder_BWT-Clustering_appendix}
    \end{subfigure}
    \begin{subfigure}[b]{0.32\textwidth}
        \centering
        \includegraphics[width=\textwidth]{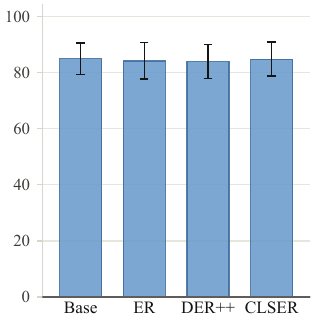}
        \caption{FWT-head}
        \label{fig:Encoder_FWT_appendix}
    \end{subfigure}
    \begin{subfigure}[b]{0.32\textwidth}
        \centering
        \includegraphics[width=\textwidth]{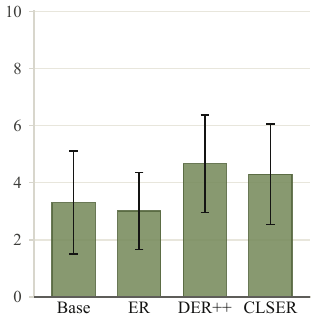}
        \caption{FWT-probe}
        \label{fig:Encoder_FWT-Prob_appendix}
    \end{subfigure}
    \begin{subfigure}[b]{0.32\textwidth}
        \centering
        \includegraphics[width=\textwidth]{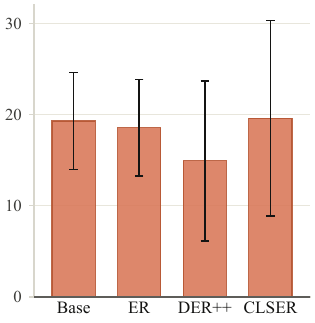}
        \caption{FWT-backbone}
        \label{fig:Encoder_FWT-Clustering_appendix}
    \end{subfigure}
    \caption{Per-method BWT and operational FWT of the discriminative backbones in the TIL scenario under the three interfaces (mean and standard deviation).}
    \label{fig:til3}
\end{figure}

We compare BWT and FWT under the three interfaces using four Incremental Learning methods across four discriminative backbones, as in the CIL scenario. As shown in Table~\ref{table:til} (and in Figure~\ref{fig:til1} for RoBERTa-base), the backbone diagnostic yields the strongest negative BWT--FWT correlation for all four backbones. Even in the TIL scenario, where a separate classifier is used for each task, the stability--plasticity pattern appears more clearly in the backbone diagnostic than in the trained head.

Observing the x-axis in Figure~\ref{fig:til_sanity}, BWT-head values are close to 0 in the TIL scenario, as separate classifiers are used for each task. However, prior research \citep{zhou2021directprobe} has suggested that classifiers can account for much of the performance, even when the model's representation is relatively weak. In comparison, BWT-backbone spans a much broader range of values, reflecting differences in representation change across models and IL methods that are less visible through head-level accuracy. BWT-probe, as in the CIL scenario, is close to 0 for all models and methods because the probe is retrained for each evaluation.

On the y-axis of Figure~\ref{fig:til_sanity}, the operational FWT results for the three interfaces show distinct differences even in the TIL scenario. FWT-head values exceed 60 in most cases, as they are measured on heads that have no outputs for the next classes, similar to the CIL scenario. Lastly, FWT-backbone suggests performance improvements in the backbone as new tasks are learned, which aligns with the explanation provided for the CIL scenario in Section \ref{sec:4.3}.

\input{tables/til}

In Figures~\ref{fig:Encoder_BWT_appendix} and~\ref{fig:Encoder_BWT-Clustering_appendix}, ER shows the least forgetting and DER++ the most under both the trained head and the backbone diagnostic; the two interfaces differ only in the order of Base and CLSER, whose values lie within 0.35 of each other in both cases. BWT-probe (Figure~\ref{fig:Encoder_BWT-Prob_appendix}) ranks the methods differently, as expected for a probe trained separately from the backbone. In the TIL scenario, task-specific classifiers reduce global class-incremental classifier bias, but the trained head and the probe still depend on their readout mechanisms. BWT-backbone ranks the methods almost identically to BWT-head in the TIL scenario, suggesting that separability-based forgetting can be approximated without using the trained head.

Figures \ref{fig:Encoder_FWT_appendix}, \ref{fig:Encoder_FWT-Prob_appendix}, and \ref{fig:Encoder_FWT-Clustering_appendix} compare the operational FWT measurement results in the TIL scenario. In Figure \ref{fig:Encoder_FWT_appendix}, FWT-head scores all averaged above 80, showing little difference across methods. In Figure \ref{fig:Encoder_FWT-Prob_appendix}, FWT-probe recorded extremely low values, averaging below 5 for all methods, with no discernible distinction. In Figure \ref{fig:Encoder_FWT-Clustering_appendix}, similar to Figure \ref{fig:Encoder_FWT-Clustering}, the results resemble those of the CIL scenario, as the backbone undergoes the same processes except for the difference in the classifiers used for CIL and TIL. This supports our interpretation that FWT-backbone measures backbone plasticity more independently of the trained head. The corresponding correlation coefficients for all four backbones are reported in Table~\ref{table:til}.

We report TIL results for discriminative backbones. Most generative IL methods are designed for the CIL setting, and in TIL the Base method already performs well because each task has its own prediction space.

\section{Full results}
\label{appendix:E}

\subsection{Per-method results and correlations across clustering choices}
\begin{figure}[t!]
    \centering
    \begin{subfigure}[b]{0.32\textwidth}
        \centering
        \includegraphics[width=\textwidth]{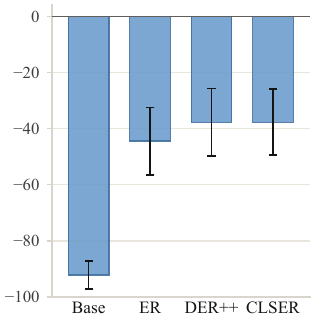}
        \caption{BWT-head}
        \label{fig:Encoder_BWT}
    \end{subfigure}    
    \begin{subfigure}[b]{0.32\textwidth}
        \centering
        \includegraphics[width=\textwidth]{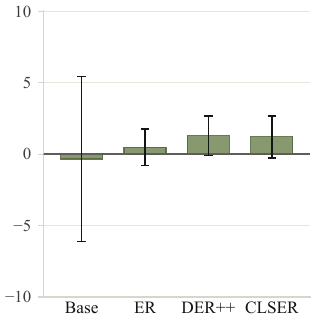}
        \caption{BWT-probe}
        \label{fig:Encoder_BWT-Prob}
    \end{subfigure}
    \begin{subfigure}[b]{0.32\textwidth}
        \centering
        \includegraphics[width=\textwidth]{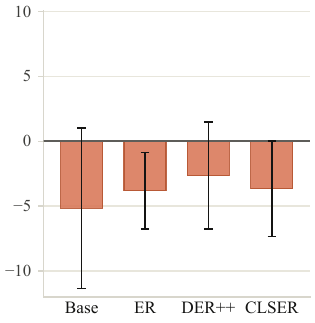}
        \caption{BWT-backbone}
        \label{fig:Encoder_BWT-Clustering}
    \end{subfigure}
    \begin{subfigure}[b]{0.32\textwidth}
        \centering
        \includegraphics[width=\textwidth]{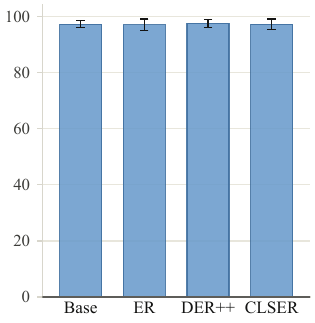}
        \caption{FWT-head}
        \label{fig:Encoder_FWT}
    \end{subfigure}
    \begin{subfigure}[b]{0.32\textwidth}
        \centering
        \includegraphics[width=\textwidth]{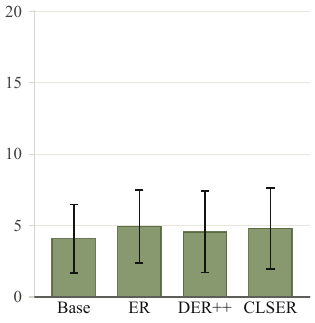}
        \caption{FWT-probe}
        \label{fig:Encoder_FWT-Prob}
    \end{subfigure}
    \begin{subfigure}[b]{0.32\textwidth}
        \centering
        \includegraphics[width=\textwidth]{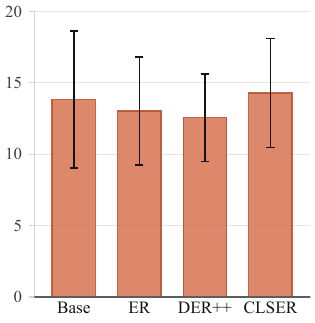}
        \caption{FWT-backbone}
        \label{fig:Encoder_FWT-Clustering}
    \end{subfigure}
    \caption{Average and standard deviation of head-level, probe-level, and backbone-level BWT/FWT for each IL method in four discriminative backbones, aggregated over the benchmark streams in Table~\ref{table:dataset}.}
    \label{fig:encoder3}
\end{figure}

Figure~\ref{fig:encoder3} shows the per-method CIL bar plots for the discriminative backbones, and Figure~\ref{fig:decoder3} in the main text shows those for the generative backbones. They show the same qualitative pattern described in Figure~\ref{fig:encoder_decocer_figure1} and Section~\ref{sec:5.1}: the trained head suggests severe forgetting, the probe suggests high recoverability, and the backbone diagnostic lies between these two views.

\input{tables/pearson}

{
\begin{figure}[t!]
    \centerline{\includegraphics[width=\columnwidth]{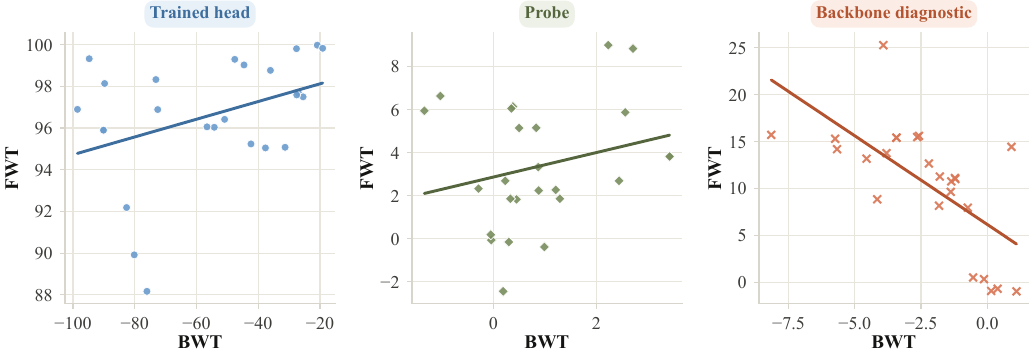}}
    \caption{BWT versus operational FWT under the three interfaces for RoBERTa-base in the CIL scenario, with least-squares fits. Correlation coefficients are reported in Tables~\ref{table:pearson} and~\ref{table:spearman}.}
    \label{fig:encoder2} 
\end{figure}
}

We conducted the same experiment for all clustering algorithms and metrics other than those analyzed in the main paper. For the discriminative backbones in Tables~\ref{table:pearson} and~\ref{table:spearman}, the BWT--FWT correlations measured through the trained head and the probe are positive (0.11 to 0.65), whereas the backbone diagnostic is negative in every configuration. This suggests that the stability--plasticity pattern is interface-dependent rather than a direct consequence of the metric formula.

The backbone diagnostic is negative regardless of whether ACC, ARI, or NMI is used. For the generative backbones, the trained-head and probe correlations are also negative ($-0.36$ to $-0.62$) but weaker than the backbone diagnostic ($-0.70$ to $-0.93$). Because the same shared-term structure applies to every interface, we use this pattern as comparative diagnostic evidence rather than as standalone proof of a universal trade-off.

{
\begin{figure}[t!]
    \centerline{\includegraphics[width=\columnwidth]{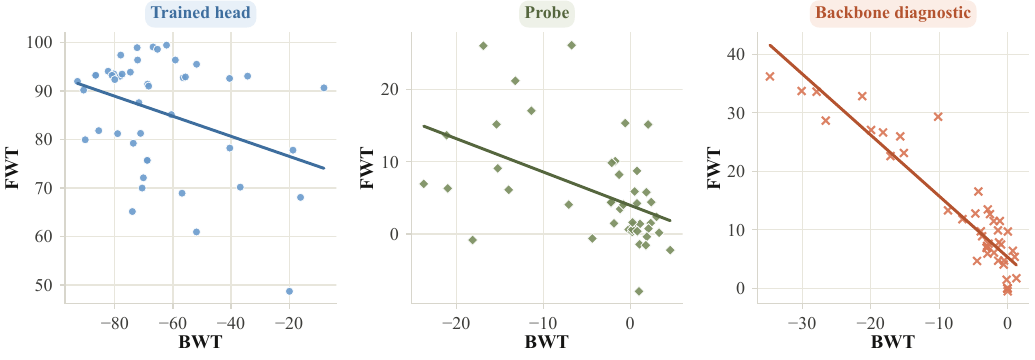}}
    \caption{BWT versus operational FWT under the three interfaces for Pythia-160M in the CIL scenario, with least-squares fits. Correlation coefficients are reported in Tables~\ref{table:pearson} and~\ref{table:spearman}.}
    \label{fig:decoder2} 
\end{figure}
}

\input{tables/Spearman}

\subsection{Additional units and clustering-free diagnostics}
\label{sec:additional_units}
The additional units comprise Llama-3.2-1B with full fine-tuning on TACRED, Banking77, CLINC150, and FewRel \(\times\) Base, DER++, and CLSER (12 units); Llama-3.2-1B with LoRA and prompt tuning on TACRED, Banking77, and CLINC150 (6 units); and Qwen3-0.6B with full fine-tuning on TACRED, Banking77, and CLINC150 \(\times\) Base, DER++, and CLSER (9 units). At every task boundary, we store the backbone features of the test sets of all tasks, including future tasks, so that every diagnostic is computed offline from the same features. NCM and \(k\)-nearest-neighbor accuracy (\(k \in \{1, 10\}\)) are measured on a fixed stratified 50/50 split of the test features, with class means or neighbors taken from one half and accuracy measured on the other; the silhouette score uses cosine distance and ground-truth labels (\(\times 100\)), and the Fisher ratio is \(100\, S_b/(S_b+S_w)\) for between-class scatter \(S_b\) and within-class scatter \(S_w\). These diagnostics use L2-normalized features. The clustering diagnostic uses Spectral Clustering without normalization, \(k=C\) clusters, seed 0, Hungarian matching, the natural class distribution, and \(n=1{,}500\) samples. Table~\ref{tab:additional_units} reports all units.

\input{tables/additional_units}

\subsection{Parameter-efficient methods}
\label{sec:peft_units}
Table~\ref{tab:peft_units} reports the six parameter-efficient units summarized in Section~\ref{sec:sp_backbone}. For prompt tuning, head accuracy remains low throughout (\(A_T\) 4.6--10.2), so its comparatively small BWT-head reflects low accuracy rather than retention.

\input{tables/peft_units}

\subsection{Full JLT results}

Tables~\ref{tab:table_main} and~\ref{tab:table_main_b} report the full per-model and per-dataset JLT results corresponding to the aggregate gap summarized in Table~\ref{table:jlt_gap_summary}.

\input{tables/table_main}
\afterpage{\clearpage\input{tables/jlt_residual}}

\subsection{Model size \& capacity}

\begin{figure}[t]
    \centering
    \begin{subfigure}[b]{0.32\textwidth}
        \centering
        \includegraphics[width=\textwidth]{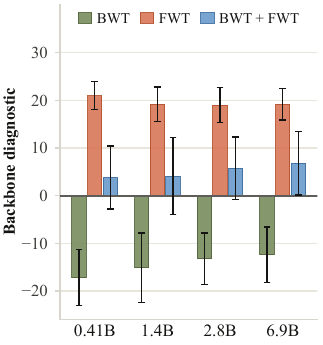}
        \caption{Pythia}
    \end{subfigure}
    \begin{subfigure}[b]{0.32\textwidth}
        \centering
        \includegraphics[width=\textwidth]{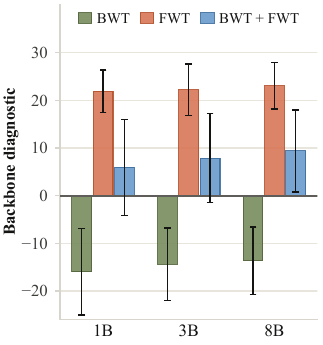}
        \caption{Llama}
    \end{subfigure}
    \begin{subfigure}[b]{0.32\textwidth}
        \centering
        \includegraphics[width=\textwidth]{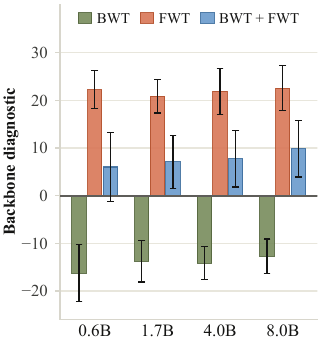}
        \caption{Qwen}
    \end{subfigure}

    \caption{BWT-backbone, FWT-backbone, and BWT-backbone + FWT-backbone results in Pythia, Llama, and Qwen, measured over five benchmark groups with the Base method.}
    \label{fig:capacity}
\end{figure}

Figure~\ref{fig:capacity} (Section~\ref{sec:scaling_observation}) reports BWT-backbone, FWT-backbone, and their sum for the Base method in three groups of representative LLMs (GPT-style Pythia, Llama, and Qwen). The sum \(BWT+FWT = a_{T,i} - a_{i-1,i}\) cancels the shared post-training term \(a_{i,i}\), so we treat it as a coupling-aware net-transfer view rather than as a primary metric. The sum increases with model size within each model family. In particular, BWT-backbone improves from the smallest to the largest model in every family, indicating a positive relationship between model size and retention. On the other hand, FWT-backbone changes little across models and sizes. The relatively stable operational FWT despite larger model sizes should be considered when designing future IL methods.

\section{LM-head adaptation details}
\label{sec:lmhead_details}

\begin{figure}
\begin{center}
\includegraphics[width=1.0\linewidth]{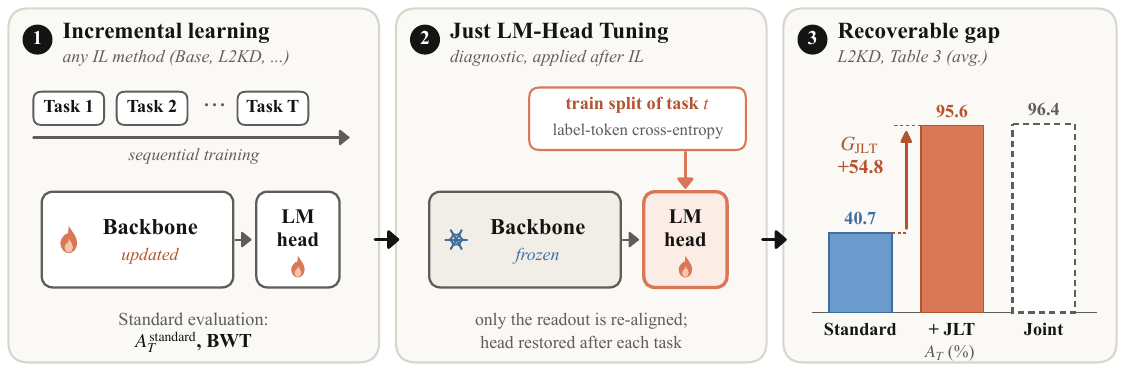}
\end{center}
\caption{Overview of JLT as a head-only diagnostic. (1) The IL process follows the basic procedure of each IL method, updating the backbone and LM head task by task. (2) After IL training, the backbone is frozen and only the LM head is re-aligned with a label-token cross-entropy objective. (3) The resulting gap $G_{\mathrm{JLT}}$ estimates how much performance is recoverable through head re-alignment; the bars show the L2KD averages from Table~\ref{table:jlt_gap_summary} together with the joint fine-tuning reference.}
\label{fig:jlt}
\end{figure}

\paragraph{Setup.}
For the JLT diagnostic, the backbone is fixed after sequential IL training and only the LM head is updated. We use two head-adaptation modes. In post-hoc JLT, the head is re-aligned once per checkpoint with a small replay buffer and no task-specific adaptation (Section~\ref{sec:posthoc_jlt}). In task-adaptive JLT, the backbone is frozen and the LM head is briefly adapted on the training split of the current task before that task is evaluated (see Section~\ref{sec:lmhead-eval-ft}). The latter is an oracle-style recovery diagnostic and is illustrated in Figure~\ref{fig:jlt}. The two variants differ in their optimization targets: task-adaptive JLT uses the first label token, whereas post-hoc JLT uses the full answer span, including the end-of-sequence token.

\paragraph{Label tokenization.}
For each instance, we tokenize the gold textual label and take the first token id as target:
\[
  y^{\mathrm{tok}} = \mathrm{Tokenizer}(\text{label})[0].
\]
Given hidden feature $\mathbf{h}\in\mathbb{R}^{d}$ from the backbone and output embedding $W\in\mathbb{R}^{V\times d}$, the vocabulary logits are
\[
  \mathbf{z} = W\,\mathbf{h} \in \mathbb{R}^{V}.
\]

\paragraph{Base LM-head loss.}
\[
  \mathcal{L}_{\mathrm{LM}} = \mathrm{CE}\!\left(\mathbf{z},\, y^{\mathrm{tok}}\right).
\]
Equivalently, for task \(t\) with \(N_t\) training examples, hidden features \(\mathbf{h}_t^i=f_T(x_t^i)\) from the frozen backbone after IL, and the label-token mapping \(\tau(y)=\mathrm{Tokenizer}(y)[0]\), the task-level head-only objective is
\[
\mathcal{L}_t
= -\frac{1}{N_t}\sum_{i=1}^{N_t}\log \softmax(W \mathbf{h}_t^i)\big[\tau(y_t^i)\big].
\]

\paragraph{Optimization.}
The LM head parameters are optimized by AdamW with learning rate $\eta$ (\(5\times10^{-4}\) by default).
During JLT, only the LM-head parameters are passed to this optimizer; the backbone receives no gradient updates.

\begin{table}[t]
\centering
\caption{Hyperparameters for LM-head adaptation.}
\label{tab:lmhead-hypers}
\small
\begin{tabular}{@{}ll@{}}
\toprule
\textbf{Hyperparameter} & \textbf{Default} \\
\midrule
LM-head learning rate & \(5\times10^{-4}\) \\
Optimizer & AdamW \\
Loss & First-label-token CE (task-adaptive); answer-token CE (post-hoc) \\
Epochs (\(E\)) & 3 \\
Batching & Same as the backbone training loop \\
\bottomrule
\end{tabular}
\end{table}

\subsection{Evaluation-time LM-head finetune}
\label{sec:lmhead-eval-ft}
Before evaluation on task $t$, the backbone is frozen and the LM head is adapted for $E$ epochs (1--3; 3 by default, Table~\ref{tab:lmhead-hypers}) on the training split of task $t$ using $\mathcal{L}_{\mathrm{LM}}$.
Evaluation itself is unchanged: the model generates the label sequence, and a prediction is counted as correct only if it exactly matches the gold label. After evaluation, LM-head weights are restored to their original state to avoid leakage across tasks. Because this protocol uses task-specific training data at evaluation time, it is reported only as a diagnostic estimate of recoverable head performance.

\subsection{Post-hoc JLT}
\label{sec:posthoc_jlt}
Post-hoc JLT is a deployable variant under the same rehearsal assumption as ER, namely that a small number of examples from previously seen tasks can be accessed. After training on task \(k\) and before evaluating checkpoint \(k\), the LM head is re-aligned once; after evaluation, the original head is restored exactly. For each seen task \(0,\dots,k\), the buffer takes as many examples as the task has classes from its shuffled training loader (batch size 4), rounded up to whole batches. This gives 8, 12, and 12 examples per task for TACRED (5 classes per task), Banking77 (11), and CLINC150 (10), and 64, 84, and 180 examples in total at the final checkpoint. Unlike task-adaptive JLT, post-hoc JLT never accesses the full training split of an evaluated task and never sees data from future tasks, so its \(A_T\), BWT, and FWT are standard CIL quantities. The buffer size is thus comparable to ER's one example per class, and the buffer is redrawn from the training loaders at each checkpoint.

Only the output embedding (LM head) is trained; the backbone is kept in evaluation mode without gradients. The loss is cross-entropy computed only at the answer-token positions (the gold label and end-of-sequence tokens), optimized with AdamW (learning rate \(5\times10^{-4}\), weight decay 0.01, no scheduler, no gradient clipping) for 3 epochs with batch size 4. Experiments use Llama-3.2-1B in bf16 with FlashAttention-2, seed 42, and 5 IL training epochs per task. Table~\ref{tab:posthoc_jlt} reports the results. Post-hoc JLT uses a separate set of Llama-3.2-1B runs, so its standard baselines differ from those in Tables~\ref{tab:table_main} and~\ref{tab:table_main_b}.

\input{tables/posthoc_jlt}

\section{Comparison with prior analyses}
\label{sec:prior_comparison}
Table~\ref{tab:prior_comparison} compares our framework with prior analyses of representation forgetting. Prior work mainly measures stability and shows that much previous-task information remains in the backbone while the head degrades; it does not quantify representation-level plasticity or compare the BWT--FWT relation across interfaces.

\input{tables/prior_comparison}

\label{sec:appendix}


\FloatBarrier

%% file: sections/impact_statement.tex
\section*{Broader impacts}

This work does not release a new pretrained model or dataset. Its positive impact is to make continual learning evaluation more transparent by separating deployed performance from recoverable representation information. By re-examining BWT and FWT through the trained head, a probe, and the backbone diagnostic, we show that transfer metrics can lead to different conclusions depending on where they are measured, which supports more careful and reproducible evaluation.

A possible negative impact is that diagnostic recoverability could be mistaken for deployable performance, leading to overoptimistic claims about model updating. We mitigate this by explicitly separating standard IL results from diagnostic JLT recoverability and by recommending that future work report the evaluation interface used for each metric.

%% file: tables/compute_resources.tex
\begin{table}[t]
\centering
\caption{Estimated compute resources for the reported experiments. Each run uses a single GPU, and eight GPUs were used in parallel. Runtimes are approximate ranges measured on our local GPUs. The number of runs counts benchmark-stream/model/method/seed configurations used for the reported results.}
\label{tab:compute_resources}

\small
\begin{tabular}{@{}llccc@{}}
\toprule
\textbf{Experiment group} & \textbf{GPU} & \textbf{Runtime per run} & \textbf{\# Runs} & \textbf{GPU-hours} \\
\midrule
Discriminative CIL & 8\(\times\) RTX A5000 (24GB) & 0.5--3 h & 240 & \(\approx\)120--720 \\
Generative CIL, small models & 8\(\times\) RTX A5000 (24GB) & 1--4 h & 720 & \(\approx\)720--2{,}880 \\
JLT diagnostic & 8\(\times\) RTX A5000 (24GB) & 0.2--1 h & 180 & \(\approx\)36--180 \\
Model-size analysis & 8\(\times\) RTX A5000 (24GB) & 2--8 h & 135 & \(\approx\)270--1{,}080 \\
\bottomrule
\end{tabular}
\end{table}

%% file: tables/dataset.tex
\begin{table}[!t]
\centering
\caption{Statistics of the incremental streams built from the sentence-level datasets used in our experiments. \textbf{\# Tasks} is the number of incremental tasks, \textbf{\# CIL classes} is the number of test-set classes in the CIL scenario, and \textbf{\# TIL classes} is the number of test-set classes evaluated for each task in the TIL scenario.}
\label{table:dataset}

\small
\setlength{\tabcolsep}{4.5pt}
\begin{tabular}{@{}llcccccc@{}}
\toprule
\textbf{Task} & \textbf{Dataset} & \textbf{\# Classes} & \textbf{\# Tasks} & \makecell{\textbf{\# CIL}\\\textbf{classes}} & \makecell{\textbf{\# TIL}\\\textbf{classes}} & \makecell{\textbf{\# Training}\\\textbf{instances}} & \makecell{\textbf{\# Test}\\\textbf{instances}} \\
\midrule
Text classification & Topic3Datasets & 25 & 5 & 25 & 5 & 75{,}000 & 46{,}000 \\
\midrule
\multirow{2}{*}{Intent classification} & CLINC150 & 150 & 15 & 150 & 10 & 15{,}000 & 4{,}500 \\
 & Banking77 & 77 & 7 & 77 & 11 & 7{,}191 & 2{,}800 \\
\midrule
\multirow{2}{*}{Relation extraction} & FewRel & 80 & 8 & 80 & 10 & 33{,}600 & 11{,}200 \\
 & TACRED & 40 & 8 & 40 & 5 & 5{,}909 & 1{,}259 \\
\bottomrule
\end{tabular}
\end{table}

%% file: tables/assets_license.tex
\begin{table}[t]
\centering
\caption{Existing datasets and pretrained model assets used in this work. We cite the original papers in the main text and summarize the public license or access terms reported by the corresponding source or model card.}
\label{tab:assets_license}

\footnotesize
\renewcommand{\arraystretch}{1.15}
\begin{tabularx}{\textwidth}{@{}>{\raggedright\arraybackslash}p{0.19\textwidth}l>{\raggedright\arraybackslash}X>{\raggedright\arraybackslash}p{0.29\textwidth}@{}}
\toprule
\textbf{Asset} & \textbf{Type} & \textbf{Version / source} & \textbf{License or access terms} \\
\midrule
Topic3Datasets (AGNews, DBPedia, Yahoo) & Dataset & Zhang et al.\ topic-classification datasets; DBpedia source data from DBpedia releases & AGNews/Yahoo: research-use terms from the original sources; DBpedia: CC-BY-SA/GFDL terms for DBpedia releases \\
CLINC150 & Dataset & UCI Machine Learning Repository, \url{https://archive.ics.uci.edu/dataset/570/clinc150} & CC BY 4.0 \\
Banking77 & Dataset & PolyAI-LDN dataset card, \url{https://huggingface.co/datasets/PolyAI/banking77} & CC BY 4.0 \\
FewRel & Dataset & THUNLP FewRel release, \url{https://huggingface.co/datasets/thunlp/few_rel} & MIT License \\
TACRED & Dataset & Stanford TACRED / LDC2018T24, \url{https://catalog.ldc.upenn.edu/LDC2018T24} & LDC User Agreement \\
BERT & Model & Google BERT model card, e.g., \url{https://huggingface.co/google-bert/bert-base-uncased} & Apache License 2.0 \\
RoBERTa & Model & FacebookAI RoBERTa model card, e.g., \url{https://huggingface.co/FacebookAI/roberta-base} & MIT License \\
Pythia / GPT-NeoX & Model & EleutherAI Pythia model cards, e.g., \url{https://huggingface.co/EleutherAI/pythia-410m-v0} & Apache License 2.0 \\
Qwen2, Qwen2.5, Qwen3 & Model & Qwen model cards, e.g., \url{https://huggingface.co/Qwen/Qwen3-0.6B} & Apache License 2.0 \\
Llama 3.1 / 3.2 & Model & Meta Llama model cards, e.g., \url{https://huggingface.co/meta-llama/Llama-3.2-3B} & Llama Community License and Acceptable Use Policy \\
\bottomrule
\end{tabularx}
\end{table}

%% file: tables/roberta_base_er.tex
\begin{table}[t]
\centering
\caption{Clustering scores of RoBERTa-base trained with ER in the CIL scenario. The last row reports the trained head's \(A_T\), the average accuracy over all tasks after learning the final task \(T\).}
\label{table3}

\small
\setlength{\tabcolsep}{5pt}
\begin{tabular}{@{}llccccc@{}}
\toprule
\textbf{Algorithm} & \textbf{Metric} & \textbf{TACRED} & \textbf{Banking77} & \textbf{CLINC150} & \textbf{FewRel} & \textbf{Topic3} \\
\midrule
K-means & ACC & 52.04 $\pm$ 1.5 & 65.23 $\pm$ 2.0 & 78.37 $\pm$ 1.8 & 62.29 $\pm$ 2.2 & 66.50 $\pm$ 1.7 \\
 & ARI & 54.90 $\pm$ 1.3 & 71.68 $\pm$ 1.7 & 83.91 $\pm$ 2.0 & 66.18 $\pm$ 1.5 & 74.55 $\pm$ 1.8 \\
 & NMI & 71.71 $\pm$ 1.4 & 80.55 $\pm$ 1.9 & 87.32 $\pm$ 2.1 & 77.91 $\pm$ 2.0 & 84.35 $\pm$ 1.6 \\
\addlinespace
GMM & ACC & 52.06 $\pm$ 1.6 & 65.77 $\pm$ 1.8 & 78.90 $\pm$ 2.0 & 61.44 $\pm$ 2.1 & 61.70 $\pm$ 1.7 \\
 & ARI & 57.62 $\pm$ 1.5 & 74.76 $\pm$ 1.9 & 81.73 $\pm$ 2.1 & 64.29 $\pm$ 1.8 & 73.53 $\pm$ 1.6 \\
 & NMI & 71.15 $\pm$ 1.3 & 83.75 $\pm$ 2.0 & 88.40 $\pm$ 2.2 & 76.90 $\pm$ 1.9 & 82.16 $\pm$ 1.5 \\
\addlinespace
Spectral & ACC & 59.82 $\pm$ 1.4 & 67.92 $\pm$ 2.0 & 80.02 $\pm$ 1.7 & 60.18 $\pm$ 1.9 & 61.80 $\pm$ 1.8 \\
 & ARI & 55.40 $\pm$ 1.5 & 72.18 $\pm$ 1.8 & 83.18 $\pm$ 2.1 & 60.40 $\pm$ 2.0 & 76.23 $\pm$ 1.6 \\
 & NMI & 70.17 $\pm$ 1.6 & 82.95 $\pm$ 1.9 & 89.47 $\pm$ 2.2 & 75.73 $\pm$ 2.0 & 84.68 $\pm$ 1.7 \\
\addlinespace
Agglomerative & ACC & 63.82 $\pm$ 1.7 & 70.81 $\pm$ 1.8 & 79.69 $\pm$ 2.1 & 63.94 $\pm$ 1.9 & 67.28 $\pm$ 1.6 \\
 & ARI & 61.30 $\pm$ 1.4 & 75.57 $\pm$ 2.0 & 84.34 $\pm$ 2.2 & 67.95 $\pm$ 1.8 & 77.96 $\pm$ 1.7 \\
 & NMI & 72.20 $\pm$ 1.5 & 85.12 $\pm$ 2.1 & 90.57 $\pm$ 2.3 & 78.55 $\pm$ 2.0 & 84.56 $\pm$ 1.8 \\
\addlinespace
Deep Clustering & ACC & 59.58 $\pm$ 1.6 & 67.36 $\pm$ 1.9 & 80.45 $\pm$ 2.1 & 63.18 $\pm$ 1.7 & 61.44 $\pm$ 1.8 \\
 & ARI & 61.62 $\pm$ 1.5 & 74.27 $\pm$ 1.8 & 84.02 $\pm$ 2.0 & 66.50 $\pm$ 1.6 & 73.54 $\pm$ 1.9 \\
 & NMI & 70.51 $\pm$ 1.4 & 83.44 $\pm$ 2.0 & 88.79 $\pm$ 2.3 & 78.11 $\pm$ 2.1 & 83.66 $\pm$ 1.7 \\
\midrule
Trained head & $A_{T}$ & 49.19 $\pm$ 1.3 & 52.79 $\pm$ 1.8 & 66.69 $\pm$ 2.0 & 42.54 $\pm$ 1.9 & 73.23 $\pm$ 1.6 \\
\bottomrule
\end{tabular}
\end{table}

%% file: tables/pythia_410m_lamol_g.tex
\begin{table}[t]
\centering
\caption{Clustering scores of Pythia-410M trained with LAMOL\_g in the CIL scenario. The last row reports the trained head's \(A_T\), the average accuracy over all tasks after learning the final task \(T\).}
\label{table5}

\small
\setlength{\tabcolsep}{5pt}
\begin{tabular}{@{}llccccc@{}}
\toprule
\textbf{Algorithm} & \textbf{Metric} & \textbf{TACRED} & \textbf{Banking77} & \textbf{CLINC150} & \textbf{FewRel} & \textbf{Topic3} \\
\midrule
K-means & ACC & 27.36 $\pm$ 3.2 & 67.13 $\pm$ 2.8 & 71.39 $\pm$ 4.5 & 57.23 $\pm$ 3.1 & 63.23 $\pm$ 2.4 \\
 & ARI & 24.66 $\pm$ 2.6 & 71.23 $\pm$ 3.1 & 76.24 $\pm$ 4.2 & 62.12 $\pm$ 2.9 & 68.22 $\pm$ 2.7 \\
 & NMI & 45.67 $\pm$ 2.3 & 81.93 $\pm$ 3.0 & 85.25 $\pm$ 3.8 & 74.33 $\pm$ 4.5 & 80.91 $\pm$ 3.1 \\
\addlinespace
GMM & ACC & 27.77 $\pm$ 2.8 & 65.38 $\pm$ 3.4 & 71.89 $\pm$ 3.7 & 58.36 $\pm$ 4.2 & 59.27 $\pm$ 2.5 \\
 & ARI & 23.97 $\pm$ 3.1 & 70.46 $\pm$ 2.6 & 77.82 $\pm$ 4.1 & 61.99 $\pm$ 3.7 & 72.23 $\pm$ 3.0 \\
 & NMI & 46.03 $\pm$ 3.4 & 82.23 $\pm$ 2.9 & 84.91 $\pm$ 3.3 & 73.10 $\pm$ 4.4 & 81.85 $\pm$ 3.5 \\
\addlinespace
Spectral & ACC & 28.74 $\pm$ 4.1 & 66.40 $\pm$ 2.5 & 71.44 $\pm$ 3.6 & 55.13 $\pm$ 3.9 & 48.80 $\pm$ 2.8 \\
 & ARI & 21.46 $\pm$ 3.7 & 69.45 $\pm$ 2.7 & 76.61 $\pm$ 4.4 & 56.08 $\pm$ 4.3 & 73.90 $\pm$ 3.1 \\
 & NMI & 46.38 $\pm$ 3.2 & 82.42 $\pm$ 2.8 & 86.06 $\pm$ 4.1 & 72.96 $\pm$ 3.5 & 84.48 $\pm$ 2.7 \\
\addlinespace
Agglomerative & ACC & 28.30 $\pm$ 3.1 & 68.12 $\pm$ 2.6 & 72.16 $\pm$ 3.8 & 61.79 $\pm$ 3.7 & 76.09 $\pm$ 3.9 \\
 & ARI & 24.00 $\pm$ 2.7 & 71.44 $\pm$ 3.5 & 78.11 $\pm$ 3.3 & 65.40 $\pm$ 3.1 & 80.59 $\pm$ 4.2 \\
 & NMI & 46.99 $\pm$ 2.9 & 83.86 $\pm$ 2.4 & 87.65 $\pm$ 4.0 & 76.31 $\pm$ 4.1 & 83.22 $\pm$ 3.5 \\
\addlinespace
Deep Clustering & ACC & 28.12 $\pm$ 2.8 & 66.32 $\pm$ 3.6 & 71.23 $\pm$ 4.2 & 56.12 $\pm$ 3.5 & 65.42 $\pm$ 4.3 \\
 & ARI & 24.34 $\pm$ 3.3 & 69.23 $\pm$ 3.4 & 77.24 $\pm$ 3.7 & 63.78 $\pm$ 4.1 & 71.10 $\pm$ 3.6 \\
 & NMI & 46.23 $\pm$ 2.6 & 83.23 $\pm$ 3.5 & 86.11 $\pm$ 3.9 & 74.60 $\pm$ 4.2 & 82.12 $\pm$ 3.3 \\
\midrule
Trained head & $A_{T}$ & 28.95 $\pm$ 3.2 & 51.95 $\pm$ 4.3 & 34.38 $\pm$ 3.1 & 23.09 $\pm$ 4.2 & 74.65 $\pm$ 3.4 \\
\bottomrule
\end{tabular}
\end{table}

%% file: tables/sensitivity.tex
\begin{table}[t]
\centering
\caption{Sensitivity of the clustering-based backbone diagnostic: BWT--FWT Pearson correlation over the 27 additional units under 25 settings, each recomputed offline from the same saved features. The default uses Spectral Clustering without normalization, \(k=C\) clusters, seed 0, Hungarian matching, the natural class distribution, and \(n=1{,}500\) samples; each row in the upper block changes one factor, and the lower block crosses three algorithms with four normalizations.}
\label{tab:sensitivity}

\small
\begin{tabular}{@{}llccc@{}}
\toprule
\textbf{Factor} & \textbf{Setting} & \textbf{ACC} & \textbf{ARI} & \textbf{NMI} \\
\midrule
Default & Spectral, no norm., \(k{=}C\), seed 0, Hungarian & $-$0.67 & $-$0.70 & $-$0.58 \\
\midrule
Normalization & L2 & $-$0.67 & $-$0.66 & $-$0.52 \\
 & Standardization & $-$0.64 & $-$0.65 & $-$0.50 \\
 & Whitening & $-$0.63 & $-$0.57 & $-$0.51 \\
\addlinespace
Number of clusters & \(k=0.5C\) & $-$0.82 & $-$0.63 & $-$0.48 \\
 & \(k=2C\) & $-$0.81 & $-$0.82 & $-$0.60 \\
\addlinespace
Seed & 1 & $-$0.67 & $-$0.71 & $-$0.57 \\
 & 2 & $-$0.70 & $-$0.75 & $-$0.61 \\
\addlinespace
Label matching & Majority vote & $-$0.75 & $-$0.70 & $-$0.58 \\
Class balance & Balanced subsampling & $-$0.62 & $-$0.68 & $-$0.50 \\
\addlinespace
Sample size & \(n=750\) & $-$0.65 & $-$0.68 & $-$0.42 \\
 & \(n=3{,}000\) & $-$0.63 & $-$0.64 & $-$0.60 \\
 & Full test set & $-$0.66 & $-$0.62 & $-$0.61 \\
\midrule
\multicolumn{5}{@{}l}{\textit{Algorithm \(\times\) normalization}} \\
K-means & None & $-$0.60 & $-$0.57 & $-$0.51 \\
 & L2 & $-$0.57 & $-$0.61 & $-$0.55 \\
 & Standardization & $-$0.62 & $-$0.60 & $-$0.51 \\
 & Whitening & $-$0.61 & $-$0.39 & $-$0.47 \\
\addlinespace
GMM & None & $-$0.58 & $-$0.55 & $-$0.47 \\
 & L2 & $-$0.56 & $-$0.60 & $-$0.52 \\
 & Standardization & $-$0.49 & $-$0.58 & $-$0.59 \\
 & Whitening & $-$0.46 & $-$0.26 & $-$0.38 \\
\addlinespace
Agglomerative & None & $-$0.62 & $-$0.60 & $-$0.58 \\
 & L2 & $-$0.66 & $-$0.38 & $-$0.56 \\
 & Standardization & $-$0.60 & $-$0.58 & $-$0.57 \\
 & Whitening & $-$0.62 & $-$0.38 & $-$0.32 \\
\bottomrule
\end{tabular}
\end{table}

%% file: tables/til.tex
\begin{table}[t]
\centering
\caption{Pearson correlation coefficient between BWT and operational FWT for each interface and clustering metric in the TIL scenario. The results aggregate the benchmark streams in Table~\ref{table:dataset} for four methods: Base, ER, DER++, and CLSER.}
\label{table:til}
\small
\setlength{\tabcolsep}{3.4pt}
\begin{tabular}{@{}llcccc@{}}
\toprule
 & & \multicolumn{4}{c}{\textbf{Discriminative backbones}} \\
\cmidrule(l){3-6}
\textbf{Interface} & \textbf{Metric} & \makecell{BERT-base} & \makecell{BERT-large} & \makecell{RoBERTa-base} & \makecell{RoBERTa-large} \\
\midrule
Trained head & ACC & $-$0.46 & $-$0.61 & $-$0.51 & $-$0.63 \\
Probe & ACC & $-$0.55 & $-$0.50 & $-$0.47 & $-$0.62 \\
\midrule
\multicolumn{6}{@{}l}{\textit{Backbone diagnostic}} \\
K-means & ACC & $-$0.84 & $-$0.80 & $-$0.87 & $-$0.81 \\
 & ARI & $-$0.88 & $-$0.81 & $-$0.80 & $-$0.87 \\
 & NMI & $-$0.88 & $-$0.83 & $-$0.84 & $-$0.82 \\
\addlinespace
GMM & ACC & $-$0.80 & $-$0.88 & $-$0.87 & $-$0.78 \\
 & ARI & $-$0.89 & $-$0.88 & $-$0.80 & $-$0.85 \\
 & NMI & $-$0.88 & $-$0.85 & $-$0.82 & $-$0.84 \\
\addlinespace
Spectral & ACC & $-$0.81 & $-$0.87 & $-$0.83 & $-$0.80 \\
 & ARI & $-$0.87 & $-$0.85 & $-$0.89 & $-$0.87 \\
 & NMI & $-$0.85 & $-$0.82 & $-$0.84 & $-$0.83 \\
\addlinespace
Agglomerative & ACC & $-$0.87 & $-$0.85 & $-$0.80 & $-$0.89 \\
 & ARI & $-$0.91 & $-$0.88 & $-$0.82 & $-$0.88 \\
 & NMI & $-$0.89 & $-$0.84 & $-$0.86 & $-$0.80 \\
\addlinespace
Deep Clustering & ACC & $-$0.81 & $-$0.79 & $-$0.88 & $-$0.83 \\
 & ARI & $-$0.88 & $-$0.82 & $-$0.84 & $-$0.82 \\
 & NMI & $-$0.88 & $-$0.85 & $-$0.81 & $-$0.89 \\
\bottomrule
\end{tabular}
\end{table}

%% file: tables/pearson.tex
\begin{table}[t]
\centering
\caption{Pearson correlation coefficient between BWT and operational FWT for each interface and clustering metric in the CIL scenario.}
\label{table:pearson}
\footnotesize
\setlength{\tabcolsep}{3.4pt}
\begin{tabular}{@{}llcccccccccc@{}}
\toprule
 & & \multicolumn{4}{c}{\textbf{Discriminative backbones}} & \multicolumn{6}{c}{\textbf{Generative backbones}} \\
\cmidrule(lr){3-6}\cmidrule(l){7-12}
\textbf{Interface} & \textbf{Metric} & \makecell{BERT\\base} & \makecell{BERT\\large} & \makecell{RoBERTa\\base} & \makecell{RoBERTa\\large} & \makecell{Pythia\\70M} & \makecell{Pythia\\160M} & \makecell{Pythia\\410M} & \makecell{Qwen2\\0.5B} & \makecell{Qwen2.5\\0.5B} & \makecell{Qwen3\\0.6B} \\
\midrule
Trained head & ACC & 0.43 & 0.14 & 0.38 & 0.24 & $-$0.50 & $-$0.36 & $-$0.37 & $-$0.52 & $-$0.36 & $-$0.38 \\
Probe & ACC & 0.65 & 0.11 & 0.25 & 0.11 & $-$0.42 & $-$0.48 & $-$0.62 & $-$0.43 & $-$0.48 & $-$0.62 \\
\midrule
\multicolumn{12}{@{}l}{\textit{Backbone diagnostic}} \\
K-means & ACC & $-$0.63 & $-$0.56 & $-$0.48 & $-$0.52 & $-$0.85 & $-$0.87 & $-$0.85 & $-$0.85 & $-$0.87 & $-$0.86 \\
 & ARI & $-$0.66 & $-$0.51 & $-$0.53 & $-$0.54 & $-$0.80 & $-$0.81 & $-$0.82 & $-$0.80 & $-$0.81 & $-$0.82 \\
 & NMI & $-$0.68 & $-$0.61 & $-$0.58 & $-$0.68 & $-$0.82 & $-$0.91 & $-$0.89 & $-$0.82 & $-$0.91 & $-$0.90 \\
\addlinespace
GMM & ACC & $-$0.52 & $-$0.57 & $-$0.50 & $-$0.43 & $-$0.77 & $-$0.75 & $-$0.77 & $-$0.77 & $-$0.75 & $-$0.78 \\
 & ARI & $-$0.68 & $-$0.69 & $-$0.62 & $-$0.51 & $-$0.81 & $-$0.80 & $-$0.84 & $-$0.81 & $-$0.80 & $-$0.85 \\
 & NMI & $-$0.69 & $-$0.70 & $-$0.61 & $-$0.61 & $-$0.79 & $-$0.83 & $-$0.84 & $-$0.79 & $-$0.84 & $-$0.85 \\
\addlinespace
Spectral & ACC & $-$0.69 & $-$0.61 & $-$0.62 & $-$0.60 & $-$0.81 & $-$0.83 & $-$0.87 & $-$0.81 & $-$0.83 & $-$0.86 \\
 & ARI & $-$0.68 & $-$0.62 & $-$0.68 & $-$0.65 & $-$0.82 & $-$0.85 & $-$0.81 & $-$0.83 & $-$0.85 & $-$0.82 \\
 & NMI & $-$0.72 & $-$0.64 & $-$0.66 & $-$0.69 & $-$0.82 & $-$0.89 & $-$0.87 & $-$0.83 & $-$0.89 & $-$0.87 \\
\addlinespace
Agglomerative & ACC & $-$0.47 & $-$0.51 & $-$0.61 & $-$0.46 & $-$0.77 & $-$0.75 & $-$0.70 & $-$0.77 & $-$0.75 & $-$0.70 \\
 & ARI & $-$0.74 & $-$0.69 & $-$0.60 & $-$0.74 & $-$0.87 & $-$0.81 & $-$0.89 & $-$0.86 & $-$0.82 & $-$0.89 \\
 & NMI & $-$0.91 & $-$0.71 & $-$0.73 & $-$0.75 & $-$0.88 & $-$0.93 & $-$0.87 & $-$0.88 & $-$0.92 & $-$0.88 \\
\addlinespace
Deep Clustering & ACC & $-$0.50 & $-$0.49 & $-$0.62 & $-$0.59 & $-$0.79 & $-$0.82 & $-$0.78 & $-$0.80 & $-$0.82 & $-$0.88 \\
 & ARI & $-$0.47 & $-$0.52 & $-$0.59 & $-$0.48 & $-$0.84 & $-$0.81 & $-$0.80 & $-$0.84 & $-$0.81 & $-$0.81 \\
 & NMI & $-$0.63 & $-$0.64 & $-$0.67 & $-$0.62 & $-$0.87 & $-$0.83 & $-$0.88 & $-$0.87 & $-$0.83 & $-$0.89 \\
\bottomrule
\end{tabular}
\end{table}

%% file: tables/Spearman.tex
\begin{table}[t]
\centering
\caption{Spearman correlation coefficient between BWT and operational FWT for each interface and clustering metric in the CIL scenario.}
\label{table:spearman}
\footnotesize
\setlength{\tabcolsep}{3.4pt}
\begin{tabular}{@{}llcccccccccc@{}}
\toprule
 & & \multicolumn{4}{c}{\textbf{Discriminative backbones}} & \multicolumn{6}{c}{\textbf{Generative backbones}} \\
\cmidrule(lr){3-6}\cmidrule(l){7-12}
\textbf{Interface} & \textbf{Metric} & \makecell{BERT\\base} & \makecell{BERT\\large} & \makecell{RoBERTa\\base} & \makecell{RoBERTa\\large} & \makecell{Pythia\\70M} & \makecell{Pythia\\160M} & \makecell{Pythia\\410M} & \makecell{Qwen2\\0.5B} & \makecell{Qwen2.5\\0.5B} & \makecell{Qwen3\\0.6B} \\
\midrule
Trained head & ACC & 0.42 & 0.13 & 0.38 & 0.24 & $-$0.51 & $-$0.37 & $-$0.37 & $-$0.53 & $-$0.38 & $-$0.37 \\
Probe & ACC & 0.64 & 0.11 & 0.24 & 0.10 & $-$0.42 & $-$0.47 & $-$0.63 & $-$0.43 & $-$0.49 & $-$0.63 \\
\midrule
\multicolumn{12}{@{}l}{\textit{Backbone diagnostic}} \\
K-means & ACC & $-$0.62 & $-$0.56 & $-$0.49 & $-$0.52 & $-$0.84 & $-$0.87 & $-$0.85 & $-$0.84 & $-$0.86 & $-$0.85 \\
 & ARI & $-$0.66 & $-$0.52 & $-$0.53 & $-$0.54 & $-$0.81 & $-$0.80 & $-$0.83 & $-$0.81 & $-$0.81 & $-$0.83 \\
 & NMI & $-$0.65 & $-$0.61 & $-$0.58 & $-$0.67 & $-$0.82 & $-$0.92 & $-$0.89 & $-$0.82 & $-$0.91 & $-$0.89 \\
\addlinespace
GMM & ACC & $-$0.51 & $-$0.58 & $-$0.49 & $-$0.43 & $-$0.76 & $-$0.74 & $-$0.78 & $-$0.77 & $-$0.75 & $-$0.78 \\
 & ARI & $-$0.67 & $-$0.69 & $-$0.62 & $-$0.52 & $-$0.82 & $-$0.81 & $-$0.85 & $-$0.82 & $-$0.81 & $-$0.86 \\
 & NMI & $-$0.55 & $-$0.59 & $-$0.61 & $-$0.61 & $-$0.79 & $-$0.84 & $-$0.84 & $-$0.79 & $-$0.83 & $-$0.84 \\
\addlinespace
Spectral & ACC & $-$0.69 & $-$0.61 & $-$0.61 & $-$0.60 & $-$0.82 & $-$0.83 & $-$0.86 & $-$0.82 & $-$0.83 & $-$0.86 \\
 & ARI & $-$0.68 & $-$0.63 & $-$0.68 & $-$0.65 & $-$0.82 & $-$0.85 & $-$0.81 & $-$0.82 & $-$0.85 & $-$0.81 \\
 & NMI & $-$0.62 & $-$0.64 & $-$0.65 & $-$0.61 & $-$0.82 & $-$0.88 & $-$0.88 & $-$0.82 & $-$0.88 & $-$0.87 \\
\addlinespace
Agglomerative & ACC & $-$0.47 & $-$0.51 & $-$0.61 & $-$0.46 & $-$0.78 & $-$0.74 & $-$0.70 & $-$0.79 & $-$0.74 & $-$0.71 \\
 & ARI & $-$0.74 & $-$0.69 & $-$0.60 & $-$0.73 & $-$0.86 & $-$0.81 & $-$0.89 & $-$0.87 & $-$0.81 & $-$0.89 \\
 & NMI & $-$0.90 & $-$0.71 & $-$0.73 & $-$0.62 & $-$0.88 & $-$0.92 & $-$0.88 & $-$0.88 & $-$0.92 & $-$0.88 \\
\addlinespace
Deep Clustering & ACC & $-$0.50 & $-$0.49 & $-$0.61 & $-$0.58 & $-$0.80 & $-$0.83 & $-$0.89 & $-$0.81 & $-$0.83 & $-$0.89 \\
 & ARI & $-$0.48 & $-$0.53 & $-$0.58 & $-$0.49 & $-$0.85 & $-$0.82 & $-$0.79 & $-$0.85 & $-$0.82 & $-$0.80 \\
 & NMI & $-$0.52 & $-$0.54 & $-$0.60 & $-$0.63 & $-$0.87 & $-$0.72 & $-$0.79 & $-$0.88 & $-$0.72 & $-$0.79 \\
\bottomrule
\end{tabular}
\end{table}

%% file: tables/additional_units.tex
\begin{table}[p]
\centering
\caption{Additional units: BWT (top) and operational FWT (bottom) under each interface. Llama: Llama-3.2-1B; Qwen: Qwen3-0.6B. Spectral is the clustering-based diagnostic in its default setting; NCM, kNN, silhouette (Silh.), and Fisher ratio are clustering-free. \(A_T\) is the trained head's final average accuracy. The last row reports the BWT--FWT Pearson correlation over the 27 units.}
\label{tab:additional_units}

\footnotesize
\setlength{\tabcolsep}{4.2pt}
\renewcommand{\arraystretch}{0.97}
\begin{tabular}{@{}lllcccccccc@{}}
\toprule
 & & & & & \multicolumn{6}{c}{\textbf{Backbone diagnostic}} \\
\cmidrule(l){6-11}
\textbf{Backbone} & \textbf{Dataset} & \textbf{Method} & \(A_T\) & \textbf{Head} & \textbf{Spectral} & \textbf{NCM} & \textbf{kNN-1} & \textbf{kNN-10} & \textbf{Silh.} & \textbf{Fisher} \\
\midrule
\multicolumn{11}{@{}l}{\textit{BWT}} \\
Llama & Banking77 & LoRA & 14.90 & $-$86.06 & $-$1.91 & $-$3.94 & $-$1.06 & $-$2.95 & $-$7.05 & $-$3.39 \\
Llama & Banking77 & Prompt tuning & 5.23 & $-$18.64 & 4.11 & 1.06 & 2.58 & 0.61 & $-$2.83 & $-$4.37 \\
Llama & CLINC150 & LoRA & 12.02 & $-$88.95 & $-$8.90 & $-$3.81 & $-$3.33 & $-$4.19 & $-$8.14 & $-$5.73 \\
Llama & CLINC150 & Prompt tuning & 10.20 & $-$49.12 & $-$5.04 & $-$1.33 & $-$0.76 & $-$1.19 & $-$2.76 & $-$2.38 \\
Llama & TACRED & LoRA & 16.00 & $-$76.54 & $-$10.74 & $-$6.32 & $-$1.07 & $-$6.41 & $-$4.08 & $-$6.92 \\
Llama & TACRED & Prompt tuning & 4.59 & $-$50.31 & $-$4.38 & $-$6.66 & $-$4.67 & $-$4.01 & $-$1.62 & $-$1.65 \\
Llama & Banking77 & CLSER & 15.23 & $-$91.29 & $-$1.02 & $-$9.39 & $-$0.83 & $-$8.18 & $-$11.60 & $-$2.56 \\
Llama & Banking77 & DER++ & 14.58 & $-$93.26 & $-$7.22 & $-$12.58 & $-$4.55 & $-$8.71 & $-$15.58 & $-$7.17 \\
Llama & Banking77 & Base & 26.98 & $-$82.35 & $-$3.71 & $-$2.27 & $-$0.76 & $-$1.82 & $-$6.88 & $-$4.64 \\
Llama & CLINC150 & CLSER & 6.09 & $-$77.93 & $-$7.09 & $-$11.00 & $-$3.14 & $-$5.71 & $-$7.65 & 0.53 \\
Llama & CLINC150 & DER++ & 6.09 & $-$80.88 & $-$7.11 & $-$12.71 & $-$1.57 & $-$8.29 & $-$10.61 & $-$2.09 \\
Llama & CLINC150 & Base & 47.56 & $-$55.36 & $-$5.63 & $-$1.57 & $-$1.76 & $-$2.57 & $-$4.63 & $-$2.43 \\
Llama & FewRel & CLSER & 46.31 & $-$58.14 & $-$0.32 & $-$1.92 & 1.37 & $-$0.37 & $-$5.11 & $-$4.88 \\
Llama & FewRel & DER++ & 48.40 & $-$54.96 & 1.97 & 0.51 & 1.80 & 0.96 & $-$1.19 & $-$0.46 \\
Llama & FewRel & Base & 37.80 & $-$68.38 & 0.41 & $-$2.88 & $-$0.12 & $-$1.49 & $-$8.19 & $-$6.53 \\
Llama & TACRED & CLSER & 25.99 & $-$80.69 & 3.05 & 3.29 & 2.63 & 1.41 & 0.58 & $-$1.76 \\
Llama & TACRED & DER++ & 35.44 & $-$69.18 & $-$3.75 & $-$2.08 & $-$2.78 & $-$3.58 & $-$5.25 & $-$3.78 \\
Llama & TACRED & Base & 33.68 & $-$72.07 & $-$14.57 & $-$5.99 & $-$9.95 & $-$9.57 & $-$10.84 & $-$11.32 \\
Qwen & Banking77 & CLSER & 12.63 & $-$88.71 & $-$17.59 & $-$36.36 & $-$21.97 & $-$24.70 & $-$42.12 & $-$26.20 \\
Qwen & Banking77 & DER++ & 13.47 & $-$87.39 & $-$12.53 & $-$25.53 & $-$10.68 & $-$18.41 & $-$31.87 & $-$16.51 \\
Qwen & Banking77 & Base & 24.54 & $-$84.89 & $-$0.95 & $-$1.89 & $-$0.30 & $-$2.80 & $-$12.86 & $-$9.40 \\
Qwen & CLINC150 & CLSER & 6.44 & $-$93.95 & $-$22.51 & $-$36.43 & $-$17.57 & $-$26.19 & $-$44.92 & $-$21.96 \\
Qwen & CLINC150 & DER++ & 6.42 & $-$92.74 & $-$24.38 & $-$35.24 & $-$17.71 & $-$24.90 & $-$42.16 & $-$21.32 \\
Qwen & CLINC150 & Base & 45.51 & $-$57.48 & $-$4.40 & $-$2.86 & $-$2.33 & $-$4.14 & $-$17.62 & $-$14.12 \\
Qwen & TACRED & CLSER & 16.35 & $-$90.20 & $-$6.36 & $-$21.86 & $-$9.64 & $-$16.01 & $-$10.88 & $-$6.44 \\
Qwen & TACRED & DER++ & 17.70 & $-$88.97 & $-$1.87 & $-$8.88 & $-$7.82 & $-$4.47 & $-$5.18 & $-$9.25 \\
Qwen & TACRED & Base & 21.07 & $-$85.17 & $-$9.35 & $-$15.91 & $-$5.73 & $-$13.01 & $-$9.34 & $-$10.97 \\
\midrule
\multicolumn{11}{@{}l}{\textit{Operational FWT}} \\
Llama & Banking77 & LoRA & -- & 87.54 & 2.70 & 6.29 & 1.52 & 2.65 & 6.66 & 9.05 \\
Llama & Banking77 & Prompt tuning & -- & 17.65 & $-$0.73 & 2.35 & 3.26 & 2.65 & 4.19 & 3.94 \\
Llama & CLINC150 & LoRA & -- & 90.90 & 6.53 & 2.38 & 1.62 & 1.76 & 7.30 & 5.92 \\
Llama & CLINC150 & Prompt tuning & -- & 48.31 & 7.73 & 1.71 & 1.24 & 1.14 & 4.18 & 3.46 \\
Llama & TACRED & LoRA & -- & 80.75 & 2.37 & 4.10 & 0.96 & 4.86 & 0.53 & 1.82 \\
Llama & TACRED & Prompt tuning & -- & 47.52 & 2.76 & 4.29 & 0.92 & 3.67 & 0.74 & $-$0.40 \\
Llama & Banking77 & CLSER & -- & 93.45 & 16.89 & 19.24 & 8.48 & 14.09 & 30.18 & 20.05 \\
Llama & Banking77 & DER++ & -- & 94.70 & 14.13 & 20.45 & 6.74 & 14.02 & 31.21 & 22.27 \\
Llama & Banking77 & Base & -- & 95.80 & 6.06 & 6.67 & 3.41 & 4.32 & 14.16 & 9.76 \\
Llama & CLINC150 & CLSER & -- & 83.81 & 14.31 & 31.05 & 18.14 & 22.67 & 43.08 & 29.73 \\
Llama & CLINC150 & DER++ & -- & 86.76 & 13.60 & 28.81 & 15.90 & 21.38 & 43.99 & 26.20 \\
Llama & CLINC150 & Base & -- & 89.86 & 6.33 & 2.81 & 2.38 & 4.19 & 11.78 & 8.53 \\
Llama & FewRel & CLSER & -- & 91.59 & 9.38 & 8.73 & 4.63 & 6.27 & 12.57 & 10.43 \\
Llama & FewRel & DER++ & -- & 91.37 & 2.61 & 5.43 & 2.59 & 4.61 & 6.50 & 5.02 \\
Llama & FewRel & Base & -- & 90.05 & 9.66 & 9.37 & 4.73 & 5.80 & 16.27 & 11.74 \\
Llama & TACRED & CLSER & -- & 95.84 & $-$1.35 & 3.46 & 5.24 & 4.21 & 4.22 & 6.11 \\
Llama & TACRED & DER++ & -- & 95.43 & $-$0.38 & 2.35 & 3.22 & 2.64 & 1.14 & 5.03 \\
Llama & TACRED & Base & -- & 96.22 & 4.36 & 3.61 & $-$1.40 & 3.42 & 4.37 & 6.49 \\
Qwen & Banking77 & CLSER & -- & 88.90 & 13.30 & 37.05 & 15.68 & 21.67 & 45.13 & 22.80 \\
Qwen & Banking77 & DER++ & -- & 88.56 & 11.01 & 29.92 & 13.26 & 20.30 & 42.48 & 23.19 \\
Qwen & Banking77 & Base & -- & 96.32 & 8.65 & 7.35 & 1.82 & 8.18 & 20.10 & 14.07 \\
Qwen & CLINC150 & CLSER & -- & 93.95 & 25.56 & 47.76 & 28.67 & 34.81 & 60.27 & 32.60 \\
Qwen & CLINC150 & DER++ & -- & 92.71 & 27.16 & 44.38 & 25.62 & 33.24 & 55.08 & 27.68 \\
Qwen & CLINC150 & Base & -- & 96.05 & 10.40 & 5.29 & 3.48 & 7.05 & 22.26 & 14.86 \\
Qwen & TACRED & CLSER & -- & 95.10 & 2.85 & 12.13 & 5.30 & 5.55 & 8.43 & 16.50 \\
Qwen & TACRED & DER++ & -- & 95.41 & $-$4.87 & $-$2.30 & $-$3.09 & $-$3.08 & 1.19 & 2.04 \\
Qwen & TACRED & Base & -- & 95.60 & 11.57 & 11.21 & 10.28 & 14.49 & 14.81 & 16.49 \\
\midrule
\multicolumn{4}{@{}l}{Pearson \(r\)(BWT, FWT), 27 units} & $-$0.746 & $-$0.672 & $-$0.872 & $-$0.616 & $-$0.800 & $-$0.837 & $-$0.497 \\
\bottomrule
\end{tabular}
\end{table}

%% file: tables/peft_units.tex
\begin{table}[t]
\centering
\caption{Parameter-efficient IL on Llama-3.2-1B (CIL). Spectral: clustering-based backbone diagnostic. NCM: nearest-class-mean accuracy on backbone features, evaluated on a fixed stratified 50/50 split of the test features (no cluster count, initialization, or cluster-label matching). The last row reports the BWT--FWT Pearson correlation over the six units.}
\label{tab:peft_units}

\small
\setlength{\tabcolsep}{5pt}
\begin{tabular}{@{}llccccccc@{}}
\toprule
 & & & \multicolumn{2}{c}{\textbf{Head}} & \multicolumn{2}{c}{\textbf{Spectral}} & \multicolumn{2}{c}{\textbf{NCM}} \\
\cmidrule(lr){4-5}\cmidrule(lr){6-7}\cmidrule(l){8-9}
\textbf{Dataset} & \textbf{Method} & \(A_T\) & BWT & FWT & BWT & FWT & BWT & FWT \\
\midrule
Banking77 & LoRA & 14.90 & $-$86.06 & 87.54 & $-$1.91 & 2.70 & $-$3.94 & 6.29 \\
CLINC150 & LoRA & 12.02 & $-$88.95 & 90.90 & $-$8.90 & 6.53 & $-$3.81 & 2.38 \\
TACRED & LoRA & 16.00 & $-$76.54 & 80.75 & $-$10.74 & 2.37 & $-$6.32 & 4.10 \\
Banking77 & Prompt tuning & 5.23 & $-$18.64 & 17.65 & 4.11 & $-$0.73 & 1.06 & 2.35 \\
CLINC150 & Prompt tuning & 10.20 & $-$49.12 & 48.31 & $-$5.04 & 7.73 & $-$1.33 & 1.71 \\
TACRED & Prompt tuning & 4.59 & $-$50.31 & 47.52 & $-$4.38 & 2.76 & $-$6.66 & 4.29 \\
\midrule
\multicolumn{3}{@{}l}{Pearson \(r\)(BWT, FWT), six units} & \multicolumn{2}{c}{$-$0.998} & \multicolumn{2}{c}{$-$0.580} & \multicolumn{2}{c}{$-$0.563} \\
\bottomrule
\end{tabular}
\end{table}

%% file: tables/table_main.tex
\begin{table}[p]
\centering
\caption{Performance recovered by head-only adaptation after incremental backbone training on three generative models. JLT rows should be interpreted as recoverable performance under the diagnostic head-adaptation protocol, not as direct replacements for standard CIL accuracy. Joint Fine-tuning denotes fine-tuning on all tasks simultaneously. $A_T$ is defined by Equation~\ref{equation:Acc} with $t=T$, BWT by Equation~\ref{equation:BWT}, and FWT by Equation~\ref{equation:FWT}. Shaded rows use the task-adaptive JLT protocol. This table covers TACRED, Banking77, and CLINC150; FewRel and Topic3 are reported in Table~\ref{tab:table_main_b}.}
\label{tab:table_main}
\small
\setlength{\tabcolsep}{4.5pt}
\begin{tabular}{@{}l*{9}{c}@{}}
\toprule
 & \multicolumn{3}{c}{\textbf{TACRED}} & \multicolumn{3}{c}{\textbf{Banking77}} & \multicolumn{3}{c}{\textbf{CLINC150}} \\
\cmidrule(lr){2-4}\cmidrule(lr){5-7}\cmidrule(l){8-10}
\textbf{IL method} & $A_T$ & BWT & FWT & $A_T$ & BWT & FWT & $A_T$ & BWT & FWT \\
\midrule
\multicolumn{10}{@{}l}{\textit{Pythia-410M}} \\
Joint Fine-tuning & 98.99 & -- & -- & 95.66 & -- & -- & 96.33 & -- & -- \\
SEQ* & 44.34 & 0.00 & 0.00 & 67.12 & 0.00 & 0.00 & 84.51 & 0.00 & 0.00 \\
KLDA-E & 97.20 & 0.00 & 0.00 & 93.03 & 0.00 & 0.00 & 96.62 & 0.00 & 0.00 \\
\addlinespace[1pt]
Base & 11.97 & $-$91.26 & 91.28 & 7.56 & $-$75.42 & 68.86 & 4.67 & $-$89.69 & 87.62 \\
L2KD & 24.68 & $-$79.27 & 94.39 & 52.32 & $-$35.08 & 85.72 & 26.67 & $-$73.17 & 94.74 \\
LAMOL\_KD & 32.81 & $-$69.55 & 93.68 & 49.48 & $-$54.92 & 97.08 & 42.09 & $-$60.48 & 97.50 \\
PCLL & 24.30 & $-$65.29 & 74.37 & 45.91 & $-$56.97 & 84.41 & 44.16 & $-$57.02 & 91.81 \\
\addlinespace[1pt]
\rowcolor{gray!12}Base + JLT & 98.93 & $-$0.55 & 11.38 & 75.07 & $-$19.92 & 84.58 & 72.47 & $-$25.50 & 92.29 \\
\rowcolor{gray!12}L2KD + JLT & 98.04 & $-$0.25 & 4.92 & 95.23 & $-$2.99 & 88.56 & 96.20 & $-$3.38 & 94.52 \\
\rowcolor{gray!12}LAMOL\_KD + JLT & 98.24 & 0.25 & 9.59 & 95.03 & $-$3.33 & 88.98 & 96.38 & $-$2.95 & 93.40 \\
\rowcolor{gray!12}PCLL + JLT & 96.70 & $-$0.58 & 10.29 & 93.31 & $-$3.60 & 87.48 & 96.27 & $-$3.29 & 91.36 \\
\midrule
\multicolumn{10}{@{}l}{\textit{Llama-3.2-1B}} \\
Joint Fine-tuning & 98.33 & -- & -- & 96.26 & -- & -- & 97.33 & -- & -- \\
SEQ* & 45.82 & 0.00 & 0.00 & 67.48 & 0.00 & 0.00 & 83.72 & 0.00 & 0.00 \\
KLDA-E & 97.36 & 0.00 & 0.00 & 93.27 & 0.00 & 0.00 & 96.71 & 0.00 & 0.00 \\
\addlinespace[1pt]
Base & 12.23 & $-$91.60 & 92.01 & 11.56 & $-$91.10 & 89.28 & 5.89 & $-$93.21 & 92.45 \\
L2KD & 31.83 & $-$68.77 & 81.20 & 51.56 & $-$32.46 & 76.52 & 26.44 & $-$66.62 & 87.88 \\
LAMOL\_KD & 32.75 & $-$62.04 & 86.04 & 45.85 & $-$64.91 & 87.01 & 44.80 & $-$54.05 & 92.93 \\
PCLL & 31.18 & $-$77.22 & 93.48 & 49.03 & $-$67.69 & 87.40 & 40.69 & $-$61.93 & 93.45 \\
\addlinespace[1pt]
\rowcolor{gray!12}Base + JLT & 98.63 & $-$0.74 & 21.56 & 79.47 & $-$17.08 & 87.05 & 74.91 & $-$23.33 & 93.45 \\
\rowcolor{gray!12}L2KD + JLT & 97.06 & $-$1.76 & 17.26 & 96.10 & $-$2.20 & 87.46 & 96.36 & $-$2.95 & 95.02 \\
\rowcolor{gray!12}LAMOL\_KD + JLT & 97.15 & $-$0.83 & 8.44 & 95.36 & $-$2.42 & 84.13 & 87.84 & $-$8.10 & 89.05 \\
\rowcolor{gray!12}PCLL + JLT & 96.30 & $-$0.32 & 18.01 & 94.32 & $-$2.54 & 87.33 & 96.29 & $-$3.29 & 91.43 \\
\midrule
\multicolumn{10}{@{}l}{\textit{Qwen3-0.6B}} \\
Joint Fine-tuning & 98.66 & -- & -- & 93.36 & -- & -- & 96.66 & -- & -- \\
SEQ* & 44.09 & 0.00 & 0.00 & 66.84 & 0.00 & 0.00 & 82.91 & 0.00 & 0.00 \\
KLDA-E & 96.94 & 0.00 & 0.00 & 92.81 & 0.00 & 0.00 & 96.25 & 0.00 & 0.00 \\
\addlinespace[1pt]
Base & 11.30 & $-$90.94 & 90.14 & 12.73 & $-$92.39 & 91.44 & 6.60 & $-$90.57 & 90.64 \\
L2KD & 34.67 & $-$67.41 & 90.38 & 55.78 & $-$25.99 & 75.11 & 27.31 & $-$63.83 & 86.02 \\
LAMOL\_KD & 42.20 & $-$59.61 & 94.17 & 52.44 & $-$52.31 & 97.42 & 42.71 & $-$59.83 & 93.07 \\
PCLL & 49.40 & $-$55.33 & 93.50 & 52.99 & $-$50.68 & 97.78 & 43.27 & $-$59.81 & 90.10 \\
\addlinespace[1pt]
\rowcolor{gray!12}Base + JLT & 97.49 & $-$1.73 & 15.76 & 75.26 & $-$22.58 & 86.36 & 77.24 & $-$21.31 & 93.24 \\
\rowcolor{gray!12}L2KD + JLT & 98.21 & $-$0.15 & 10.69 & 94.93 & $-$3.67 & 86.40 & 95.27 & $-$4.40 & 93.12 \\
\rowcolor{gray!12}LAMOL\_KD + JLT & 97.88 & $-$0.53 & 13.95 & 92.08 & $-$6.82 & 84.89 & 90.47 & $-$9.50 & 92.07 \\
\rowcolor{gray!12}PCLL + JLT & 96.23 & $-$0.07 & 17.69 & 92.66 & $-$5.00 & 87.93 & 95.22 & $-$4.36 & 92.36 \\
\bottomrule
\end{tabular}
\end{table}

\begin{table}[p]
\centering
\caption{Performance recovered by head-only adaptation on FewRel and Topic3 (continued from Table~\ref{tab:table_main}). Shaded rows use the task-adaptive JLT protocol.}
\label{tab:table_main_b}
\small
\setlength{\tabcolsep}{4.5pt}
\begin{tabular}{@{}l*{6}{c}@{}}
\toprule
 & \multicolumn{3}{c}{\textbf{FewRel}} & \multicolumn{3}{c}{\textbf{Topic3}} \\
\cmidrule(lr){2-4}\cmidrule(l){5-7}
\textbf{IL method} & $A_T$ & BWT & FWT & $A_T$ & BWT & FWT \\
\midrule
\multicolumn{7}{@{}l}{\textit{Pythia-410M}} \\
Joint Fine-tuning & 95.35 & -- & -- & 95.50 & -- & -- \\
SEQ* & 61.99 & 0.00 & 0.00 & 70.56 & 0.00 & 0.00 \\
KLDA-E & 94.55 & 0.00 & 0.00 & 94.53 & 0.00 & 0.00 \\
\addlinespace[1pt]
Base & 6.17 & $-$80.57 & 73.59 & 20.53 & $-$91.02 & 93.08 \\
L2KD & 30.18 & $-$53.23 & 74.57 & 58.17 & 10.27 & 40.25 \\
LAMOL\_KD & 27.70 & $-$75.60 & 95.43 & 50.08 & $-$54.98 & 92.97 \\
PCLL & 29.79 & $-$68.41 & 82.84 & 55.80 & $-$42.23 & 92.66 \\
\addlinespace[1pt]
\rowcolor{gray!12}Base + JLT & 51.31 & $-$36.10 & 77.50 & 72.21 & $-$14.80 & 79.92 \\
\rowcolor{gray!12}L2KD + JLT & 95.18 & $-$1.04 & 93.52 & 93.91 & $-$0.16 & 92.95 \\
\rowcolor{gray!12}LAMOL\_KD + JLT & 95.73 & $-$0.54 & 92.82 & 94.00 & $-$0.08 & 92.99 \\
\rowcolor{gray!12}PCLL + JLT & 94.70 & $-$1.06 & 92.15 & 93.46 & $-$0.26 & 93.51 \\
\midrule
\multicolumn{7}{@{}l}{\textit{Llama-3.2-1B}} \\
Joint Fine-tuning & 95.65 & -- & -- & 95.50 & -- & -- \\
SEQ* & 62.13 & 0.00 & 0.00 & 71.64 & 0.00 & 0.00 \\
KLDA-E & 94.68 & 0.00 & 0.00 & 94.59 & 0.00 & 0.00 \\
\addlinespace[1pt]
Base & 8.42 & $-$81.39 & 77.03 & 19.50 & $-$88.74 & 97.84 \\
L2KD & 32.75 & $-$54.67 & 78.49 & 63.17 & 17.27 & 38.09 \\
LAMOL\_KD & 29.96 & $-$71.79 & 89.10 & 48.10 & $-$56.19 & 92.72 \\
PCLL & 28.46 & $-$74.38 & 93.88 & 48.84 & $-$55.42 & 92.89 \\
\addlinespace[1pt]
\rowcolor{gray!12}Base + JLT & 51.29 & $-$36.30 & 79.57 & 73.06 & $-$13.23 & 79.85 \\
\rowcolor{gray!12}L2KD + JLT & 95.35 & $-$1.00 & 94.38 & 93.82 & $-$0.16 & 92.79 \\
\rowcolor{gray!12}LAMOL\_KD + JLT & 94.72 & $-$0.99 & 92.16 & 90.68 & $-$1.03 & 89.74 \\
\rowcolor{gray!12}PCLL + JLT & 95.42 & $-$0.97 & 94.94 & 92.24 & $-$0.10 & 91.98 \\
\midrule
\multicolumn{7}{@{}l}{\textit{Qwen3-0.6B}} \\
Joint Fine-tuning & 95.65 & -- & -- & 96.50 & -- & -- \\
SEQ* & 60.77 & 0.00 & 0.00 & 70.41 & 0.00 & 0.00 \\
KLDA-E & 94.33 & 0.00 & 0.00 & 94.21 & 0.00 & 0.00 \\
\addlinespace[1pt]
Base & 6.98 & $-$79.11 & 73.01 & 19.99 & $-$89.91 & 89.94 \\
L2KD & 32.90 & $-$58.13 & 73.79 & 62.32 & 9.90 & 44.41 \\
LAMOL\_KD & 26.45 & $-$79.12 & 95.30 & 42.09 & $-$60.48 & 93.50 \\
PCLL & 31.29 & $-$74.50 & 96.14 & 49.40 & $-$55.33 & 92.50 \\
\addlinespace[1pt]
\rowcolor{gray!12}Base + JLT & 52.55 & $-$35.27 & 78.87 & 71.03 & $-$18.32 & 82.20 \\
\rowcolor{gray!12}L2KD + JLT & 95.58 & $-$1.15 & 94.98 & 92.16 & $-$1.21 & 91.32 \\
\rowcolor{gray!12}LAMOL\_KD + JLT & 95.01 & $-$1.87 & 94.67 & 91.47 & $-$1.53 & 90.21 \\
\rowcolor{gray!12}PCLL + JLT & 94.96 & $-$1.13 & 96.55 & 89.56 & $-$1.75 & 90.11 \\
\bottomrule
\end{tabular}
\end{table}

%% file: tables/jlt_residual.tex
\begin{table}[t]
\centering
\caption{Stream-level view of the JLT decomposition, computed from Tables~\ref{tab:table_main} and~\ref{tab:table_main_b}. \textbf{Top:} residual gap \(A^{\mathrm{joint}}_T - A^{\mathrm{JLT}}_T\) per benchmark stream (negative values: JLT slightly exceeds the joint reference). \textbf{Bottom:} mean change in operational FWT after task-adaptive head re-alignment (JLT minus standard), averaged over the three backbones.}
\label{tab:jlt_residual}

\small
\begin{tabular}{@{}llccccc@{}}
\toprule
\textbf{IL method} & \textbf{Backbone} & \textbf{TACRED} & \textbf{Banking77} & \textbf{CLINC150} & \textbf{FewRel} & \textbf{Topic3} \\
\midrule
Base & Pythia-410M & 0.06 & 20.59 & 23.86 & 44.04 & 23.29 \\
 & Llama-3.2-1B & $-$0.30 & 16.79 & 22.42 & 44.36 & 22.44 \\
 & Qwen3-0.6B & 1.17 & 18.10 & 19.42 & 43.10 & 25.47 \\
\addlinespace
L2KD & Pythia-410M & 0.95 & 0.43 & 0.13 & 0.17 & 1.59 \\
 & Llama-3.2-1B & 1.27 & 0.16 & 0.97 & 0.30 & 1.68 \\
 & Qwen3-0.6B & 0.45 & $-$1.57 & 1.39 & 0.07 & 4.34 \\
\addlinespace
LAMOL\_KD & Pythia-410M & 0.75 & 0.63 & $-$0.05 & $-$0.38 & 1.50 \\
 & Llama-3.2-1B & 1.18 & 0.90 & 9.49 & 0.93 & 4.82 \\
 & Qwen3-0.6B & 0.78 & 1.28 & 6.19 & 0.64 & 5.03 \\
\addlinespace
PCLL & Pythia-410M & 2.29 & 2.35 & 0.06 & 0.65 & 2.04 \\
 & Llama-3.2-1B & 2.03 & 1.94 & 1.04 & 0.23 & 3.26 \\
 & Qwen3-0.6B & 2.43 & 0.70 & 1.44 & 0.69 & 6.94 \\
\bottomrule
\end{tabular}

\vspace{8pt}
\begin{tabular}{@{}lcc@{}}
\toprule
 & \multicolumn{2}{c}{\textbf{Mean \(\Delta\)FWT (JLT \(-\) standard)}} \\
\cmidrule(l){2-3}
\textbf{IL method} & \textbf{TACRED + FewRel} & \textbf{Banking77 + CLINC150 + Topic3} \\
\midrule
Base & $-$35.40 & $-$2.47 \\
L2KD & $-$29.51 & +21.49 \\
LAMOL\_KD & $-$40.35 & $-$4.30 \\
PCLL & $-$34.10 & $-$1.06 \\
\bottomrule
\end{tabular}
\end{table}

%% file: tables/posthoc_jlt.tex
\begin{table}[t]
\centering
\caption{Deployable post-hoc JLT on Llama-3.2-1B. The head is re-aligned once per checkpoint with a small replay buffer of roughly one example per class and restored after evaluation; no task-specific adaptation is performed. Numbers are standard CIL metrics.}
\label{tab:posthoc_jlt}

\small
\setlength{\tabcolsep}{5pt}
\begin{tabular}{@{}llccccccc@{}}
\toprule
 & & \multicolumn{2}{c}{\textbf{\boldmath\(A_T\)}} & \multicolumn{2}{c}{\textbf{BWT}} & \multicolumn{3}{c}{\textbf{FWT}} \\
\cmidrule(lr){3-4}\cmidrule(lr){5-6}\cmidrule(l){7-9}
\textbf{Dataset} & \textbf{Method} & Standard & +JLT & Standard & +JLT & Standard & +JLT & \(\Delta\) \\
\midrule
Banking77 & Base & 26.98 & 50.55 & $-$82.35 & $-$26.29 & 95.80 & 71.29 & $-$24.51 \\
CLINC150 & Base & 47.56 & 67.49 & $-$55.36 & $-$17.98 & 89.86 & 79.12 & $-$10.74 \\
TACRED & DER++ & 35.44 & 59.13 & $-$69.18 & $-$13.34 & 95.43 & 64.71 & $-$30.72 \\
TACRED & Base & 33.68 & 61.69 & $-$72.07 & $-$10.79 & 96.22 & 67.46 & $-$28.76 \\
\midrule
\multicolumn{2}{@{}l}{Average} & 35.92 & 59.72 & $-$69.74 & $-$17.10 & 94.33 & 70.65 & $-$23.68 \\
\bottomrule
\end{tabular}
\end{table}

%% file: tables/prior_comparison.tex
\begin{table}[H]
\centering
\caption{Comparison with prior analyses of forgetting at the representation level.}
\label{tab:prior_comparison}

\small
\setlength{\tabcolsep}{3pt}
\begin{tabular}{@{}llccccc@{}}
\toprule
\textbf{Work} & \textbf{Interfaces} & \textbf{BWT} & \makecell{\textbf{FWT /}\\\textbf{plasticity}} & \makecell{\textbf{Classifier-}\\\textbf{free}} & \makecell{\textbf{Generative}\\\textbf{LM head}} & \makecell{\textbf{Interface-wise}\\\textbf{BWT--FWT}} \\
\midrule
\citet{davari2022probing} & Head, probe & \checkmark & -- & -- & -- & -- \\
\citet{zhou2021directprobe} & DirectProbe & -- & -- & Partial & -- & -- \\
\citet{seq*} & Head, probe & \checkmark & -- & -- & Partial & -- \\
\citet{zheng2025spurious} & Head & \checkmark & -- & -- & \checkmark & -- \\
\citet{chen2023forgetting} & Head & \checkmark & Head only & -- & -- & -- \\
\midrule
Ours & \makecell[l]{Head, probe,\\backbone, JLT} & \checkmark & \makecell{Representation\\level} & \checkmark & \checkmark & \checkmark \\
\bottomrule
\end{tabular}
\end{table}

%% file: checklist.tex
\section*{NeurIPS Paper Checklist}

\begin{enumerate}

\item {\bf Claims}
    \item[] Question: Do the main claims made in the abstract and introduction accurately reflect the paper's contributions and scope?
    \item[] Answer: \answerYes{}
    \item[] Justification: The abstract and the introduction (Section~\ref{sec:intro}) state the paper's scope and main contributions: treating BWT/FWT as interface-dependent quantities, separating deployed performance, recoverability, and representation separability, and using JLT only as a diagnostic head-realignment intervention.

\item {\bf Limitations}
    \item[] Question: Does the paper discuss the limitations of the work performed by the authors?
    \item[] Answer: \answerYes{}
    \item[] Justification: The paper includes a Limitations section at the beginning of the appendix, discussing the scope of the classifier-free backbone diagnostics, resource limits for larger generative models, and future work.

\item {\bf Theory assumptions and proofs}
    \item[] Question: For each theoretical result, does the paper provide the full set of assumptions and a complete (and correct) proof?
    \item[] Answer: \answerNA{}
    \item[] Justification: The paper does not present formal theoretical theorems or proofs; its claims are empirical and are supported by experiments and analysis.

    \item {\bf Experimental result reproducibility}
    \item[] Question: Does the paper fully disclose all the information needed to reproduce the main experimental results of the paper to the extent that it affects the main claims and/or conclusions of the paper (regardless of whether the code and data are provided or not)?
    \item[] Answer: \answerYes{}
    \item[] Justification: The paper describes the problem setup and metrics in Section~\ref{sec:2.1} and Section~\ref{sec:2.2}, datasets in Table~\ref{table:dataset}, model settings in Appendix~\ref{Appendix_A}, and implementation details in Appendix~\ref{Appendix_A.2}.

\item {\bf Open access to data and code}
    \item[] Question: Does the paper provide open access to the data and code, with sufficient instructions to faithfully reproduce the main experimental results, as described in supplemental material?
    \item[] Answer: \answerYes{}
    \item[] Justification: Source code and reproduction instructions are publicly available at \url{https://github.com/kkt94/RevisitingCL}.

\item {\bf Experimental setting/details}
    \item[] Question: Does the paper specify all the training and test details (e.g., data splits, hyperparameters, how they were chosen, type of optimizer) necessary to understand the results?
    \item[] Answer: \answerYes{}
    \item[] Justification: Section~\ref{sec:experimental_setup} and Appendix~\ref{Appendix_A} specify the task setup, datasets, backbone models, prompts for generative backbones, optimizer choices, learning rates, sequence lengths, and training epochs.

\item {\bf Experiment statistical significance}
    \item[] Question: Does the paper report error bars suitably and correctly defined or other appropriate information about the statistical significance of the experiments?
    \item[] Answer: \answerYes{}
    \item[] Justification: Unless otherwise noted, experiments are repeated three times. Appendix~\ref{Appendix_A.2} states that reported means average over three random seeds and that figure error bars denote one standard deviation across seeds. Correlation analyses report coefficients over dataset--method units. Table~\ref{table:interface_summary} reports mean, standard deviation, range, and sign pattern across diagnostic configurations, and explicitly treats clustering variants as non-independent robustness configurations rather than as independent statistical samples.

\item {\bf Experiments compute resources}
    \item[] Question: For each experiment, does the paper provide sufficient information on the computer resources (type of compute workers, memory, time of execution) needed to reproduce the experiments?
    \item[] Answer: \answerYes{}
    \item[] Justification: Appendix~\ref{Appendix_A.2} reports the GPU types and memory used for the experiments, and Table~\ref{tab:compute_resources} provides estimated runtime ranges, run counts, and aggregate GPU-hour ranges for each reported experiment group.
    
\item {\bf Code of ethics}
    \item[] Question: Does the research conducted in the paper conform, in every respect, with the NeurIPS Code of Ethics \url{https://neurips.cc/public/EthicsGuidelines}?
    \item[] Answer: \answerYes{}
    \item[] Justification: The research uses existing datasets and pretrained models for empirical evaluation and does not involve new data collection, surveillance, or human-subject experiments.

\item {\bf Broader impacts}
    \item[] Question: Does the paper discuss both potential positive societal impacts and negative societal impacts of the work performed?
    \item[] Answer: \answerYes{}
    \item[] Justification: The Broader Impacts section discusses the positive impact of more transparent continual learning evaluation and the negative risk that diagnostic recoverability could be mistaken for deployable performance.
    
\item {\bf Safeguards}
    \item[] Question: Does the paper describe safeguards that have been put in place for responsible release of data or models that have a high risk for misuse (e.g., pre-trained language models, image generators, or scraped datasets)?
    \item[] Answer: \answerNA{}
    \item[] Justification: The paper does not release a new pretrained language model, scraped dataset, or other high-risk asset; it uses existing pretrained models and public or licensed benchmarks for evaluation.

\item {\bf Licenses for existing assets}
    \item[] Question: Are the creators or original owners of assets (e.g., code, data, models), used in the paper, properly credited and are the license and terms of use explicitly mentioned and properly respected?
    \item[] Answer: \answerYes{}
    \item[] Justification: The paper cites the existing datasets, models, and baseline methods used in the experiments, and Table~\ref{tab:assets_license} summarizes the public license or access terms for each dataset group and pretrained model family used in the study.

\item {\bf New assets}
    \item[] Question: Are new assets introduced in the paper well documented and is the documentation provided alongside the assets?
    \item[] Answer: \answerYes{}
    \item[] Justification: The paper introduces code for the proposed evaluation and JLT procedure, which is documented and released at \url{https://github.com/kkt94/RevisitingCL} under the MIT License; no new dataset or pretrained model checkpoint is introduced.

\item {\bf Crowdsourcing and research with human subjects}
    \item[] Question: For crowdsourcing experiments and research with human subjects, does the paper include the full text of instructions given to participants and screenshots, if applicable, as well as details about compensation (if any)? 
    \item[] Answer: \answerNA{}
    \item[] Justification: The paper does not conduct new crowdsourcing or human-subject studies; it evaluates methods on existing benchmark datasets.

\item {\bf Institutional review board (IRB) approvals or equivalent for research with human subjects}
    \item[] Question: Does the paper describe potential risks incurred by study participants, whether such risks were disclosed to the subjects, and whether Institutional Review Board (IRB) approvals (or an equivalent approval/review based on the requirements of your country or institution) were obtained?
    \item[] Answer: \answerNA{}
    \item[] Justification: The paper does not involve new crowdsourcing, participant recruitment, or human-subject experiments, so IRB approval is not applicable.

\item {\bf Declaration of LLM usage}
    \item[] Question: Does the paper describe the usage of LLMs if it is an important, original, or non-standard component of the core methods in this research? Note that if the LLM is used only for writing, editing, or formatting purposes and does \emph{not} impact the core methodology, scientific rigor, or originality of the research, declaration is not required.
    \item[] Answer: \answerYes{}
    \item[] Justification: The paper makes essential use of language models as experimental backbones, including Pythia, Llama, and Qwen models, and describes their use in Section~\ref{sec:experimental_setup} and Appendix~\ref{Appendix_A.1}.

\end{enumerate}